%% file: main.tex
\documentclass[10pt,twocolumn,letterpaper]{article}

\usepackage[preprint]{iccv}      
\usepackage{mathtools}
\usepackage{ulem}
\usepackage{url}            
\usepackage{booktabs}       
\usepackage{amsfonts}       
\usepackage{nicefrac}       
\usepackage{microtype}      
\usepackage{xcolor}         
\usepackage{amsmath} 
\usepackage{multirow}
\usepackage{colortbl}
\usepackage{graphicx}
\usepackage{float}
\usepackage{colortbl}

\input{preamble}

\definecolor{iccvblue}{rgb}{0.21,0.49,0.74}
\usepackage[pagebackref,breaklinks,colorlinks,allcolors=iccvblue]{hyperref}

\def\paperID{1234} 
\def\confName{ICCV}
\def\confYear{2025}

\title{\method{}: Intention-Driven Visual Perception via Reinforced Reasoning}

\author{
   Zhangquan Chen$^{1}$\thanks{\*The work was conducted during the internship of Zhangquan Chen
(czq23@mails.tsinghua.edu.cn) at Microsoft Research Asia.} $\quad{}$  Xufang Luo$^{2}$\thanks{\*Corresponding author (xufluo@microsoft.com)}
   $\quad{}$  Dongsheng Li$^{2}$\\
   $^{1}$Tsinghua University, Beijing, China $\quad{}$ $^{2}$ Microsoft Research Asia, Shanghai, China 
}

\begin{document}
\maketitle
\input{Sec/abstract}
\input{Figs/intro}
\input{Sec/intro}
\input{Sec/related}
\input{Sec/method}

\input{Sec/exp}
\input{Sec/conclusion}
{
    \small
    \bibliographystyle{ieeenat_fullname}
    \bibliography{main}
}

\clearpage
\newpage
\appendix
\input{Sec/supmat}

\end{document}

%% file: preamble.tex
\newcommand{\method}{{\fontfamily{lmtt}\selectfont\textsc{VisRL}}\xspace}

%% file: Sec/abstract.tex
\begin{abstract}
Visual understanding is inherently intention-driven—humans selectively focus on different regions of a scene based on their goals. Recent advances in large multimodal models (LMMs) enable flexible expression of such intentions through natural language, allowing queries to guide visual reasoning processes. Frameworks like Visual Chain-of-Thought have demonstrated the benefit of incorporating explicit reasoning steps, where the model predicts a focus region before answering a query. However, existing approaches rely heavily on supervised training with annotated intermediate bounding boxes, which severely limits scalability due to the combinatorial explosion of intention-region pairs. To overcome this limitation, we propose VisRL, the first framework that applies reinforcement learning (RL) to the problem of intention-driven visual perception. VisRL optimizes the entire visual reasoning process using only reward signals. By treating intermediate focus selection as an internal decision optimized through trial-and-error, our method eliminates the need for costly region annotations while aligning more closely with how humans learn to perceive the world. Extensive experiments across multiple benchmarks show that VisRL consistently outperforms strong baselines, demonstrating both its effectiveness and its strong generalization across different LMMs. Our code is available at \url{https://github.com/zhangquanchen/VisRL}.

\end{abstract}

%% file: Figs/intro.tex
\begin{figure}[!htbp]
    \centering
    \includegraphics[width=0.45\textwidth]{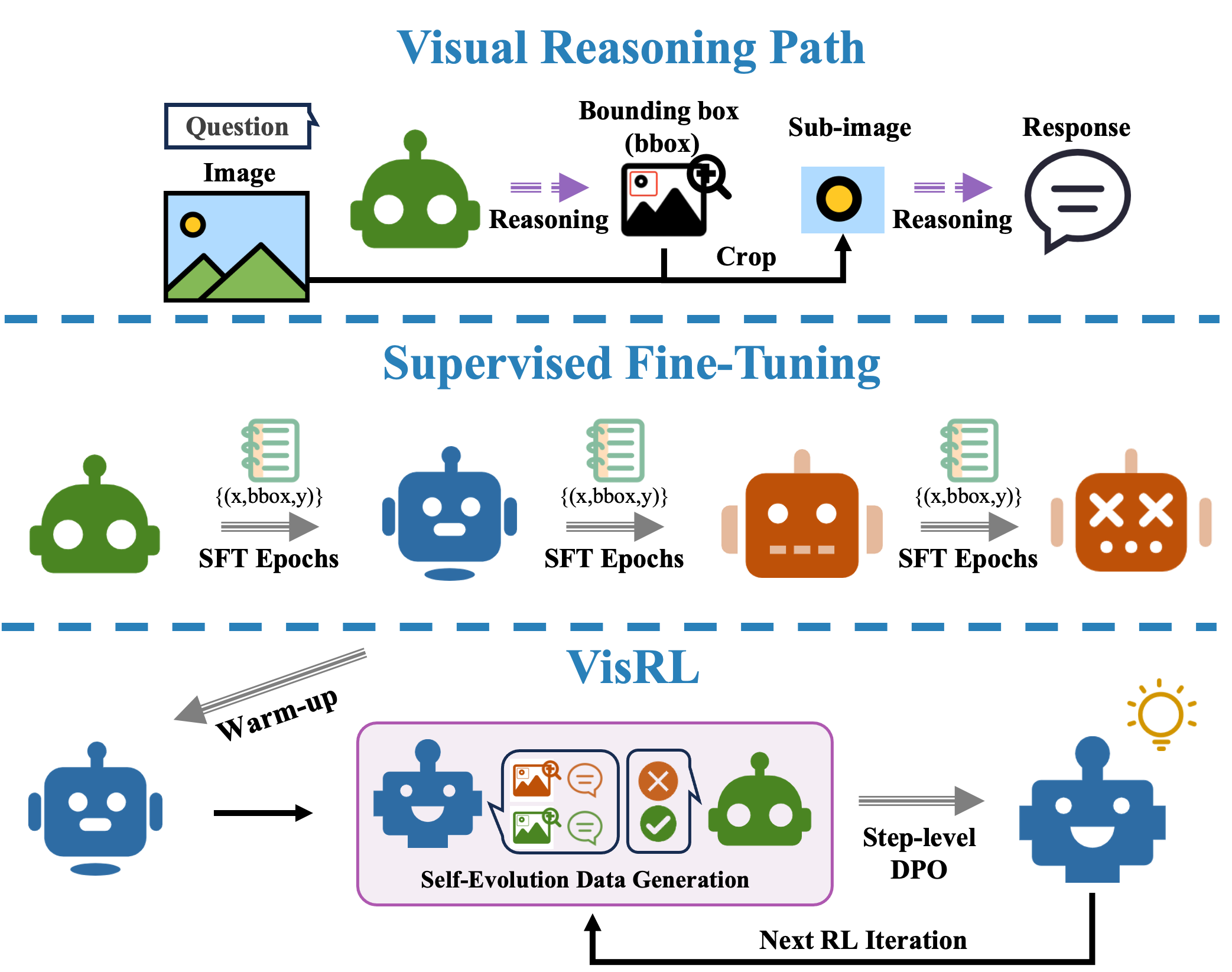}
    \vspace{-0.5em}
    \caption{Illustration of using RL to optimize the visual reasoning process. SFT trains with densely annotated training data for several epochs. \method leverages self-generated data and self-provided rewards to iteratively update the model using step-level DPO. This RL process removes the need for bounding box annotations, enabling a more human-like, intention-driven visual perception.}\label{fig:intro}
    \vspace{-2.3em}
\end{figure}


%% file: Sec/intro.tex
\section{Introduction}
\label{sec:intro}
\input{Figs/pipeline}
Visual understanding is a fundamental problem in computer vision, enabling machines to interpret and interact with their surroundings~\cite{wolfe2017five,intro2:guo2016deep,intro2:palmeri2004visual}. Traditional methods of visual perception often process entire scenes uniformly~\cite{intro:chen2024sharegpt4v,intro:chen2025sharegpt4video,intro:jiao2025lumen,intro:shao2024lmdrive,intro:zhang2024eventhallusion}, without considering the intent behind a given task. However, human perception is inherently intention-driven — people focus on different aspects of a scene depending on their goals. For example, when entering a room, a person searching for a television remote will scan tables and couches, while someone checking the time will look at the walls for a clock. This context-dependent approach to visual perception suggests that intelligent models should also adapt their focus based on the task at hand. This leads to the problem of intention-driven visual perception, where the goal is to dynamically determine the most relevant regions of an image based on a given query or task~\cite{rela:virreq}.

With the advent of large multimodal models, the intention in perception tasks can now be expressed in a highly flexible way, which is natural language. Common LMMs, such as LLaVA~\cite{mllm:liu2023llava} and Qwen-VL~\cite{mllm:wang2024qwen2}, first encode visual signals into tokens, and then both visual and text tokens are jointly processed by the large language models to produce final outputs. Despite showing powerful ability in many tasks~\cite{video1,video2,video3, fg-l1,fg-l2,fg-l3,chen2024three1}, this kind of one-pass end-to-end LMMs still suffer from hallucinations~\cite{gunjal2024detecting} and do not explicitly address the intention-driven visual perception problem. Further extending this paradigm, instead of treating the entire process as a single black-box inference, recent works have proposed frameworks like Visual Chain-of-Thought (Visual CoT~\cite{shao2024visual}). These methods introduce an explicit reasoning step where the model first predicts a bounding box representing the critical region to focus on, crops the image to extract this region, and then feeds the cropped visual input back to the multimodal model. By conditioning the final answer on both the query and the selected focus area, this approach not only improves interpretability but also helps leverage the multi-turn in-context learning capabilities of the underlying LMM.

However, despite the advantages brought by methods like Visual CoT, these approaches also impose extremely high requirements on training data. Existing methods rely on supervised learning to teach models to produce intermediate reasoning steps. Specifically, for each intention or query, the training process requires corresponding bounding box annotations to guide the model in identifying the correct focus area. This dependence on exhaustive annotation is impractical, as the same image can correspond to vastly different regions depending on the request. As a result, the annotation complexity grows combinatorially with the diversity of queries, making it impossible to cover all potential cases in a scalable manner.

In this work, we propose a novel learning framework \method that optimizes the entire reasoning process based solely on rewards and feedback from the task itself, rather than dense annotations (shown in Figure~\ref{fig:intro}). Specifically, we treat the success or failure of the task as a reward signal and apply reinforcement learning (RL) to train the model. Our approach requires no bounding box annotations, thereby addressing the scalability challenge posed by annotation requirements. Furthermore, this learning paradigm aligns more closely with how humans acquire perceptual skills — humans do not learn to focus on specific regions through meticulously annotated training data for each task, but rather through trial-and-error interaction with the environment, gradually developing the ability to adaptively zoom into relevant regions. By leveraging this reward-driven learning strategy, our framework enables intention-driven visual perception in a more flexible, scalable, and human-like manner.

\method adopts an iterative DPO framework to complete the RL process for visual reasoning. This framework consists of multiple cycles of data generation and model optimization. During the data generation phase, we introduce a diversity controller to ensure that the generated bounding boxes cover a wide variety of potential focus regions. Additionally, we apply a filtering mechanism to select questions with appropriate difficulty levels for the current model and to identify the most effective preference pairs associated with each question. During the model optimization phase, we apply a step-level DPO algorithm, ensuring that the model learns to optimize every step of the visual reasoning process.

Our contributions can be summarized as follows.

\begin{itemize}
    \item We present \method, the first framework to apply RL to the problem of intention-driven visual perception. By addressing the data annotation bottleneck, \method establishes a learning process that is much closer to human-like visual understanding.
    \item We design a tailored data generation pipeline for \method, incorporating both a diversity controller to enhance visual exploration and further use a step-level DPO algorithm to fully exploit the collected data.
    \item Extensive experiments across multiple benchmarks demonstrate that \method consistently outperforms strong baselines. Moreover, our results show that the effectiveness of \method generalizes well across different multimodal models, highlighting its broad applicability.
\end{itemize}

%% file: Figs/pipeline.tex
\begin{figure*}[!htbp]
    \centering
    \includegraphics[width=1\textwidth]{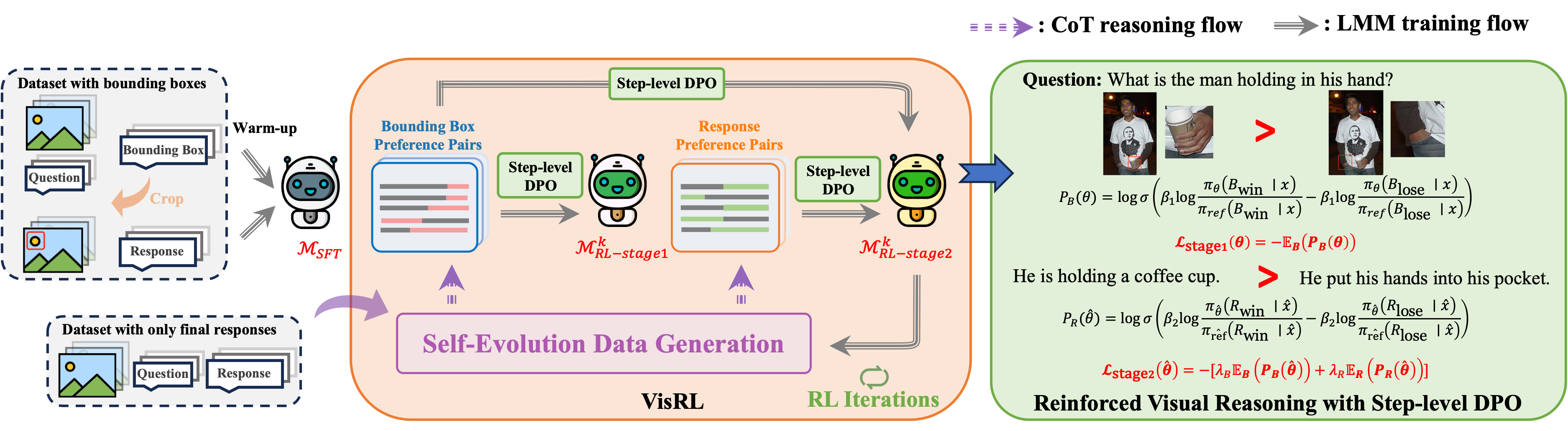}
    \vspace{-2em}
    \caption{The schematic illustration of our \method framework. \method first utilizes a small amount of data for SFT warm-up, but in the subsequent RL training phase, it can leverage large-scale data without bounding box annotations. The RL phase of \method consists of iterative cycles of data generation and optimization, and $k$ in the figure indicates the iteration index. The data generation process does not rely on external models or annotations; instead, it employs the model itself for data synthesis and scoring. The optimization step adopts step-level DPO to ensure the model learns each step of the reasoning process. In summary, \method enables intention-driven visual perception by leveraging RL to learn from task rewards without requiring annotations and external helps.}\label{fig:pipeline}
    \vspace{-1.5em}
\end{figure*}

%% file: Sec/related.tex
\section{Related Work}
\label{sec:related}

\paragraph{Multi-modal Large Language Models}
With the advancement of large language models (LLMs)~\cite{llm:radford2018improving,llm:radford2019language,llm0:floridi2020gpt,llm1:few-shot,llm10:bi2024deepseek,llm11:cai2024internlm2,llm12:jiang2024identifying,llm2:zhang2022opt,llm3:touvron2023llama,llm4:touvron2023llama,llm5:bai2023qwen,llm6:team2023internlm,llm7:almazrouei2023alghafa,llm8:glm2024chatglm,llm9:yang2023baichuan}, multi-modal large language models which integrate vision and language modalities have also experienced rapid development. This progress enables AI systems to better perceive and understand the real-world interplay between visual and textual information. Notable methods like LLaVA~\cite{mllm:liu2023llava} aligns image tokens with pre-trained LLMs by training a projector, while other approaches utilize a Q-Former~\cite{li2022blip,li2023blip} to learn image embeddings via learnable queries after extracting image features. These LMMs provide strong base models for \method, as they have the capability to process both visual and language data simultaneously, enabling the completion of the reasoning process.


\paragraph{Intention-Driven Visual Models}
Recently, several methods have attempted to enhance models' intention-driven visual perception capabilities. VisCoT~\cite{shao2024visual} employs a multi-turn interpretable processing mechanism with bounding boxes that dynamically focus on visual inputs. Similarly, SpatialCoT~\cite{rela:liu2025spatialcot} achieves spatial grounding through spatial coordinates, while SegLLM~\cite{rela:wang2024segllm} leverages mask-labeled data to enable reasoning about complex user segmentation intentions. V* (SEAL)~\cite{rela:vstar} provides an LLM-guided search mechanism for efficient visual querying. Besides, both MLLM-TPO~\cite{rela:tpo} and VisionLLM v2~\cite{rela:wu2025visionllmv2} achieve intention-driven perception by training decoders for specific downstream tasks. Additionally, MVoT~\cite{rela:mvot}introduces a text-image-text reasoning paradigm by training on interleaved data. These methods generally follow the supervised learning (SL) paradigm, therefore heavily relying on dense-labeled data (e.g., bounding boxes, spatial coordinates, masks, multi-round reasoning conversations), which constrains their ability to scale further. Instead, \method uses rewards as learning supervision and gets rid of the requirement on dense annotations. On the other hand, several methods employ tool-usage for enhancement. Specifically, AURORA~\cite{rela:aurora} leverages specialized detection models, while Plug-and-Play~\cite{rela:plugandplay} adopts a multi-agent framework. Besides, ViRReq~\cite{rela:virreq} leverages the knowledge base to decompose visual recognition. However, these approaches rely on external models or knowledge, but not focusing on enhancing the intrinsic capabilities of the models themselves, while \method tries to completing the task via learning to reasoning by the model itself. 


\paragraph{Multimodal Models with Chain-of-Thoughts} Chain-of-Thought (CoT) reasoning plays an important role for LMMs. The methods can be broadly categorized into two types. (1) Text-thought methods~\cite{rela_new:r1v, rela_new:llamavo1,rela_new:alignanything,rela:insight-v,rela:wu2024visco}
elicit textual CoT reasoning of LLMs in visual reasoning tasks by introducing text thinking tokens inspired by~\cite{rela:deepseekr1}, guiding towards the final response. Differently, \emph{\method emphasizes that visual information should be involved in the reasoning process to fully leverage the strengths of multimodal models, rather than relying solely on language tokens for reasoning}. (2) Multi-modal-thought methods involves multimodal information in the reasoning process. The Mind’s Eye~\cite{rela_new:mideye} elicits spatial reasoning of LLMs by visualizing their reasoning traces. Additionally, ~\cite{rela:liu2025spatialcot,shao2024visual,rela:wang2024segllm} first generate visual marks (e.g. bounding boxes, spatial coordinates, masks), and subsequently perform CoT reasoning based on these fine-grained visual marks. These works still uses SL for optimization while \method first explores RL in this direction. Besides, recent repositories~\cite{shen2025vlmr1, visualrft} also tried to enhance the bounding boxes (bboxes) generation through RL, but they uses ground truth bboxes to give rewards, which is still limited by dense annotation data.

%% file: Sec/method.tex
\section{Methodology}
\label{sec:method}







\paragraph{Preliminary.} Reinforcement Learning~\cite{christiano2017deep} stands out as a highly effective method for significantly bolstering the robustness, factual accuracy, and safety of large language models~\cite{ouyang2022training}. The method consists of two key training stages, namely the reward model training and the policy model training. 
To avoid this complex training pipeline, \cite{rafailov2024direct} proposed Direct Preference Optimization (DPO). DPO streamlines the process by directly leveraging pair-wise preference data to optimize the policy model with an equivalent optimization objective.
Specifically, given an input prompt $x$, and a preference data pair $(y_{win}, y_{lose})$, DPO aims to maximize the probability of the preferred output $y_{win}$ and minimize that of the undesirable output $y_{lose}$. The optimization objective is formulated as:
\begin{equation}
\mathcal{L}_{DPO}(\theta) =-\mathbb{E}_{\left(x, y_{win}, y_{lose}\right) \sim D}(P(\theta)),
\end{equation}
where $D$ is the pair-wise preference dataset, and $P_\theta$ is:
\begin{equation}
\small
P(\theta) = \log \sigma\left(\beta \log \frac{\pi_\theta\left(y_{win} \mid x\right)}{\pi_{ref}\left(y_{win} \mid x\right)}-\beta \log \frac{\pi_\theta\left(y_{lose} \mid x\right)}{\pi_{ref}\left(y_{lose} \mid x\right)}\right).
\end{equation}
$\sigma$ is the sigmoid function, $\pi_\theta(\cdot \mid x)$ is the policy model to be optimized, $\pi_{ref}(\cdot \mid x)$ is the reference model kept unchanged during training, and the hyperparameter $\beta$ serves to regulate the proximity of the policy model to the reference model.

\paragraph{Method overview.} As shown in Figure~\ref{fig:pipeline}, \method adopts DPO to optimize the entire visual reasoning process due to its simplicity. Our method leverages the final task success or failure as the outcome reward, and the grades of intermediate steps as the process reward, guiding the model to gradually refine its reasoning process through reinforcement learning. This reasoning process is divided into two steps. In the first step, the model generates a bounding box $B$ representing the focused area based on the given query or question $Q$ and the original image $I$. In the second step, the cropped region $I^s$ corresponding to the bounding box, together with the original image and the question, is fed into the multimodal model to produce the final response $R$.

In Section~\ref{sec: data-generation}, we will describe the data generation process required to support this two-stage reasoning pipeline. In Section~\ref{sec:rl}, we will introduce the optimization strategy, explaining how stepwise DPO is applied to enable the model to aquire intention-driven visual perception ability.

Notably, this data generation and optimization process will be iterated multiple times. The improved model can collect better data, and the better data will further refine the model. The initial model before RL training is denoted with $\mathcal{M}_{RL}^0$, and the model which is updated with $k$ iterations is denoted with $\mathcal{M}_{RL}^k$.

\input{Figs/data_gen}
\subsection{Data Generation}\label{sec: data-generation}
Before RL training, we use SFT as a warm-up stage. This stage changes the original model $\mathcal{M}_{org}$ into $\mathcal{M}_{SFT}$, so $\mathcal{M}_{SFT}$ is $\mathcal{M}_{RL}^0$. This stage requires bounding box annotations, so we directly use samples from VisCoT dataset~\cite{shao2024visual}. Our intent here is just to make the model have the basic ability to generate bounding boxes in specific format. Therefore, the used data amount is relatively small here (30k in ours and 438k in VisCoT). The second part of our source datasets consists of question-answer pairs from nine source datasets that span five distinct domains, with the majority of them being Visual Question Answering (VQA) and Image Captioning datasets. These are specifically used for constructing preference data for RL training, therefore requiring no bounding box annotations. We use an additional 180k data here. Then, we explain how we construct preference data based on our source datasets in details.

\noindent\textbf{Self-evolution.} The critical parts for enabling effective RL learning is generating CoT data, and providing reward signals. Unlike previous manual annotation methods, which is infeasible for providing widely enough coverage for all possible cases, or approaches that rely on more powerful external models, which is more like distillation rather than allowing the model to learn on its own, \emph{we adopt a strategy that does not depend on external capabilities—instead, leveraging the model's own self-evolution}. To achieve this, we sample from $\mathcal{M}_{SFT}$ multiple times to generate CoT data with as much diversity as possible, while use $\mathcal{M}_{org}$ to provide criticism. This approach not only showcases the intrinsic capabilities of the model, but also facilitates a more effective adjustment of the predicted probability distribution toward a stable state through self-generated data. 

\noindent\textbf{Sample generation and rewarding.} For each input question and image $(Q, I)$, we first sample from $\mathcal{M}_{SFT}$ twice to obtain the bounding boxes $B_1$ and $B_2$, respectively. Then, we introduce a diversity controller to ensure diversity of bounding boxes. Specifically, we update $B_2$ according to the Intersection over Union (IoU) and the rejecting threshold value $\mathcal{T}$ between $B_1$ and $B_2$, which is formulated as:
\begin{equation}
\hat{B_2}= \begin{cases}B_2, & \operatorname{IoU}\left(B_1, B_2\right)<\mathcal{T} \\ \mathcal{R_S}(\mathcal{U_I}(B_1)), & \operatorname{IoU}\left(B_1, B_2\right) \geq \mathcal{T}\end{cases},
\end{equation}
where $\mathcal{U_I}(B_1)$ is a set of bounding boxes that are outside of $B_1$ but within image $I$. The operator $\mathcal{R_S}(\cdot)$ represents a random selection from the set $\mathcal{U_I}(B_1)$, and have an area differing by no more than $S$ compared to $B_1$. 

Based on the bounding boxes $B_1$ and $\hat{B_2}$, we crop the sub-image $I_1^s$ and $I_2^s$ from $I$, respectively. Then, we input $(Q,I,I_1^s)$ and $(Q,I,I_2^s)$ to the model to obtain final responses $R_1$ and $R_2$, separately. At this point, we have completed the sampling of two distinct CoT reasoning paths.

Then, we should evaluate whether the paths are good or not. \method uses the original model before SFT stage $\mathcal{M}_{org}$ as a critic to score the pairs $B_1, B_2$ and $R_1, R_2$. Scores for the bounding boxes and the final responses are denoted with $s^b_1, s^b_2$ and $s^r_1, s^r_2$, respectively. For more details including prompt designs, please refer to Sec.~\ref{sec:instruction} in the Supp. Mat..

\noindent\textbf{Data filtering.} For each input question and image, repeating the above data generation process $N$ times, we will obtain a set of $2N$ candidates $P=\{p_1, p_2,...,p_{2N}\}$, where each $p$ contains original question, image, one bounding box, cropped sub-image, the final response and two scores, i.e., $p_i=(Q, I, B_i, I^s_i, R_i, s^b_i, s^r_i)$. 

Moreover, to ensure the validity of preference data pairs in the candidate set $P$, we apply filtering by setting win and lose thresholds, $\mathcal{T}^{b}_{max}$ and $\mathcal{T}^{b}_{min}$ for bounding boxes, $\mathcal{T}^{r}_{max}$ and $\mathcal{T}^{r}_{min}$ for responses:
\begin{equation}
\begin{aligned}
& P_{win}=\left\{p_i \mid s^b_i \geq \mathcal{T}_{\max }^b \text { and } s^r_i \geq \mathcal{T}_{\max }^r\right\} \\
& P_{lose}=\left\{p_i \mid s^b_i<\mathcal{T}_{\min }^b \text { and } s^r_i<\mathcal{T}_{\min }^r\right\}
\end{aligned}.
\end{equation}
For the current question and image $(Q,I)$, if the win set and the loss set are not empty, this question and image and its corresponding generated data are preserved. The intuition here is that we apply a filter to the questions in the dataset, selecting those with a difficulty level suitable for the current model. Questions that are too difficult will prevent the model from generating meaningful answers, while questions that are too easy will result in the model producing correct answers across the board. Both of these types of questions will be filtered out during the current training round. However, as the model's capabilities are updated and strengthened, less data are filtered in the next iteration, as indicated by Data Num. of VisRL-Full vs. VisRL-Full-Iter1 in Tab.~\ref{table:data}.

Then, for each preserved question, we select the most representative path from each set to obtain the win-lose preference pairs, denoted as $(p_{win},p_{lose})=(\mathcal{C}_{max}(P_{win}),\mathcal{C}_{min}(P_{lose}))$. In the case of $\mathcal{C}_{max}$, it is required that both $s^b$ and $s^r$ of $p_i$ are the maximum over the other elements in the set $P_{win}$. Conversely, $\mathcal{C}_{min}$ requires that both scores are the minimum in the set $P_{lose}$. If the condition of being simultaneously maximum/minimum on both $s^b$ and $s^r$ is not met, the data point will also be discarded. Finally, we obtain the preference dataset $D_P=\{(p_{win_1},p_{lose_1}),(p_{win_2},p_{lose_2})...\}$

\subsection{Reinforced Visual Reasoning}\label{sec:rl}


Similar with~\cite{lai2024step}, \method uses a step-level DPO method. It is divided into two stages. 
Stage 1 involves optimizing the bounding box, while stage 2 focuses on the joint optimization of the bounding box and the final response. For stage 1, given the input question-image $x = (Q,I)$, the objective is:
\begin{equation}
\mathcal{L}_{stage1}(\theta) =-\mathbb{E}_{\left(x, B_{win}, B_{lose}\right) \sim D_P}(P_B(\theta)),
\end{equation}
where each pair-wise preference paths in $D_P$ consists of bouding boxes $B_{win}$ and $B_{lose}$, the formulation of bounding box preference probability is:
\begin{equation}\label{eq:eq6}
\small
P_B(\theta) = \log \sigma\left(\beta_1 \log \frac{\pi_\theta\left(B_{win} \mid x\right)}{\pi_{ref}\left(B_{win} \mid x\right)}-\beta_1 \log \frac{\pi_\theta\left(B_{lose} \mid x\right)}{\pi_{ref}\left(B_{lose} \mid x\right)}\right).
\end{equation}
After stage1, the policy model updated from $\pi_\theta$ to $\pi_{\hat{\theta}}$, while the reference model is updated to $\pi_{\hat{ref}}$. Then, for stage2, we further consider the cropped-image from $B_{win}$ to make CoT inference, that is $\hat{x} = (Q,I,I^s_{win})$, then the formulation of response preference probability is:
\begin{equation}\label{eq:eq7}
\small
P_R(\hat{\theta}) = \log \sigma\left(\beta_2 \log \frac{\pi_{\hat{\theta}}\left(R_{win} \mid \hat{x} \right)}{\pi_{\hat{ref}}\left(R_{win} \mid \hat{x} \right)}-\beta_2 \log \frac{\pi_{\hat{\theta}}\left(R_{lose} \mid \hat{x} \right)}{\pi_{\hat{ref}}\left(R_{lose} \mid \hat{x} \right)}\right),
\end{equation}
where $R_{win}$ and $R_{lose}$ are the final preference responses in $D_P$. Based on Eq.~\ref{eq:eq6} and Eq.~\ref{eq:eq7}, the objective for jointly optimizing the bounding boxes and the responses in stage 2 can be formulated as: 
\begin{equation}
\mathcal{L}_{stage2}(\hat{\theta}) =-(\lambda_B\mathcal{L}_B(\hat{\theta})+\lambda_R\mathcal{L}_R(\hat{\theta})),
\end{equation}
where:
\begin{equation}
\mathcal{L}_B(\hat{\theta}) = \mathbb{E}_{\left(x, B_{win}, B_{lose}\right) \sim D_P}(P_B(\hat{\theta})),
\end{equation}
\begin{equation}
\mathcal{L}_R(\hat{\theta}) = \mathbb{E}_{\left(\hat{x}, R_{win}, R_{lose}\right) \sim D_P}(P_R(\hat{\theta})).
\end{equation}

%% file: Figs/data_gen.tex
\begin{figure}[!htbp]
    \centering
    \vspace{-0.8em}
    \includegraphics[width=0.45\textwidth]{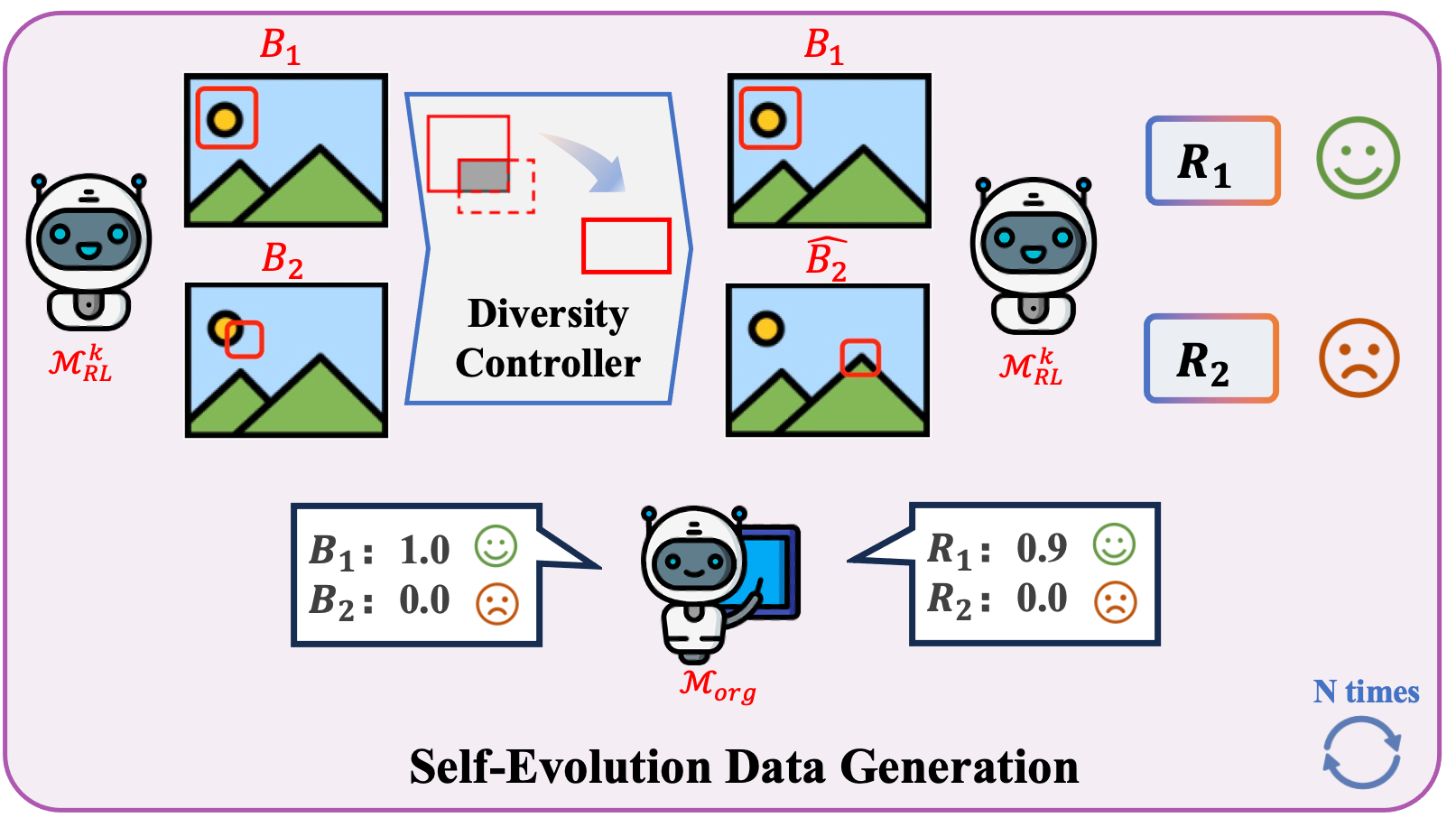}
    \vspace{-0.8em}
    \caption{The schematic illustration of our data generation pipeline. Here $\mathcal{M}_{RL}^k$ denotes the model updated with $k$ iterations of data generation and optimization, and $\mathcal{M}_{org}$ is the original model. \method{} uses $\mathcal{M}_{RL}^k$ to generate samples, and use $\mathcal{M}_{org}$ to provide rewards. Hence, different versions of a single model are used in this self-evolution data generation process, and no bounding box annotations and external models are introduced in this process.}\label{fig:data_gen}
    \vspace{-1.5em}
\end{figure}

%% file: Sec/exp.tex
\section{Experiments}\label{sec:exp}

\input{Tabs/tab_baseline}

\subsection{Comparisons with Baselines}

We evaluate our method with several state-of-the-art methods on an array of different categories as follows. More details are in Supp. Mat..
As shown in Tab.~\ref{table:baseline}, we categorize methods into three groups based on the types of LLM and vision encoder, then evaluate them on comprehensive benchmarks (MME, MMBench) as well as hallucination benchmarks (POPE). In all cases, our method achieves either the best or second-best performance, demonstrating the robust and well-rounded improvement over the baseline. In contrast, other methods exhibit performance drops on specific benchmarks (e.g., SEAL on MMBench, VisCoT on MME). We attribute this phenomenon to the limitations inherent in their training approaches -- data-driven SFT struggles to generalize effectively, while tool-usage methods suffer from intrinsic shortcomings on certain datasets. Notably, our approach achieves highly comprehensive and promising results while using only 30k dense-labeled (w. bounding boxes) samples -- 
significantly fewer than other methods, and without relying on any external capabilities. In particular, under Vicuna-7B and CLIP-ViT-L-14-336, our method outperforms VisCoT across all benchmarks -- the representative data-driven SFT approach. Specifically, our method outperforms VisCoT by \textbf{5.00\%} (1526.3 vs. 1453.6) on MME and by \textbf{1.74\%} (87.5 vs. 86.0) on the hallucination benchmark POPE. Moreover, after 1 iteration (Base Model + \method -- Iter1), \method improves performance by \textbf{1\%} to \textbf{4\%} across all benchmarks.

\subsection{Results on Visual CoT Benchmark}
In this section, we comprehensively investigate different training phases in \method across various base LMMs. We use Visual CoT benchmark here, which primarily focuses on scenarios where the LMM needs to concentrate on specific regions within a complete image.

\input{Tabs/tab_mllm}

\noindent\textbf{Settings.} 
For SFT, the objective is to regularize the model with the capability of outputting bounding box, while RL is to enhance the model's visual perception capabilities via rewards. To train the LMM with outputting bounding box in SFT phase, we add a CoT prompt (``Please first provide the bounding box coordinate of the region, then refer to the corresponding sub-image to answer the question better.") to the question, asking the model to identify the most informative region of the image. As shown in Tab.~\ref{table:sft}, we found that the model has already been capable of outputting bounding boxes under SFT with the dataset of 30k. Then, we sequentially proceed with further training using stage1 (RL1) and stage2 (RL2) as described in Sec.~\ref{sec:rl}.

\noindent\textbf{Results.} Tab.~\ref{table:mllm} indicates that after SFT, there is still a decline on some datasets even with the use of visual CoT (e.g. InfogVQA), which further corroborates the validity of our revised SFT strategy. Besides, we found that the difference in results obtained from SFT with 30k or 438k is not significant (i.e. VisCoT vs. LLaVA-1.5-7B-Ours-SFT). These suggests that: on the one hand, the improvement in model capability is more attributed to the introduction of CoT rather than the SFT memory. On the other hand, the model's capability has already achieved saturated in a data-driven SFT manner and fails to further generalize, which is also the reason why Visual CoT fails in some OOD  scenarios. However, our RL method can achieve comprehensive enhancement, with the promising improvement of up to \textbf{49.07\%} (PaliGemma2-10B: 0.377 vs. 0.562), and the minimum improvement of \textbf{23.78\%} (Llama-3.2-Vision-11B: 0.635 vs. 0.786). Meanwhile, we does not rely on a large amount of dense-labeled data  (i.e. with boundingbox annotation), but still has learned more essential visual perception. 

\noindent\textbf{Detection Ability.} Our approach is grounded in CoT for visual reasoning, thereby placing significant emphasis on the accuracy of intermediate bounding boxes. To substantiate the enhancement in bounding box precision following the implementation of our RL method, we present the detection performance in Tab.~\ref{table:detection}. Specifically, we compute the IoU 
between the predicted CoT bounding boxes and the corresponding GT bounding boxes, deeming a prediction correct if the IoU value surpasses 0.5. It is evident that, when using the same base model -- LLaVA-1.5-7B, the performance after our SFT with 30k data is somewhat inferior to that of Visual CoT which leverages 438k data. However, during the RL phase, we get rid of bounding box annotation data and exclusively utilized 180k simple question-answer pairs (further processed via our data generation pipeline in Sec.~\ref{sec: data-generation} to construct preference data), thereby achieving an accuracy improvement of \textbf{15.61\%} (0.437 vs. 0.378) over Visual CoT. Moreover, even with only RL1, we still attained an approximate \textbf{8.47\%} (0.410 vs. 0.378) improvement. Notably, on the DocVQA dataset (which was not included in the RL training phase), RL1 achieved a remarkable \textbf{39.71\%} (0.190 vs. 0.136) improvement, while RL2 accomplished an impressive \textbf{73.68\%} (0.231 vs. 0.136) improvement. We attribute these substantial gains to the robust generalization capability of our RL method.

\input{Tabs/tab_sft_bb}
\input{Tabs/tab_detection}
\input{Tabs/tab_strategy}

\subsection{Performance across Multiple RL Iterations}
Fig.~\ref{fig:chart} illustrates the performance variation curves of our method over multiple iterations on LLava-1.5-7B and Qwen2.5-VL-7B. It can be observed that after each iteration (i.e., data is regenerated and \method training is reconducted), performance improves significantly, regardless of whether only RL1 is applied or the full process of RL1+RL2 is used. The performance gains range from a minimum of \textbf{0.3\%} to a maximum of \textbf{1.8\%}. Since Qwen2.5-VL-7B is closer to the performance upper bound, its growth is relatively slower. \emph{Notably, after 4 iterations, LLava-1.5-7B even surpasses Llama-3.2-Vision-11B, as indicated by the blue curve.} This further validates the promising potential of our method and lays the experimental foundation for future optimizations in online algorithms.
\input{Figs/chart}

\subsection{Visualization}
This section presents the qualitative performance of our \method in Fig.~\ref{fig:comparison}, highlighting the accuracy of our method in identifying critical regions within images and then aid in CoT reasoning. Compared to VisCoT (\textcolor{red}{red}), our \method (\textcolor{green}{green}) demonstrates greater performance in both localizing the regions of interest and generating the final response, as evidenced by the ground truth (GT) comparison (\textcolor{blue}{blue}).
\input{Figs/comparison}

\subsection{Ablation Study}
\noindent\textbf{Different stages.}
In Tab.~\ref{table:mllm}, we have conducted the ablation study on the usage of different training stages (SFT, SFT+RL1, SFT+RL1+RL2), demonstrating a consistent performance improvement from SFT to RL1 and then to RL2. Furthermore, in Tab.~\ref{table:strategy}, we present the results of training with RL2-Only after SFT with bounding boxes. As observed, applying RL2-Only training already yields a promising improvement compared to SFT-then-CoT (0.759 vs. 0.730). However, it still results in significant performance drops on certain benchmarks compared to the original model -- Qwen2.5-VL-7B (e.g., DocVQA, DUDE, InfogVQA). This suggests that directly optimizing both bounding-box detection and response generation in a joint manner poses certain challenges. Therefore, introducing RL1 first to optimize bounding-box detection separately is necessary.

\noindent\textbf{Different training strategy.}
We attempted to directly learn the final response in Tab.~\ref{table:strategy}, i.e., without the CoT setting. In this experiment, we used the same 30k dataset $D_P$. Specifically, SFT was trained solely on the chosen response (w/o bounding box); DPO~\cite{rafailov2024direct} was trained on the win/lose preference response; Kahneman-Tversky Optimisation (KTO)~\cite{kto} assigned "true" and "false" labels to "win" and "lose" response, respectively; and Proximal Policy Optimization (PPO)~\cite{ppo} sampled 15k preference data to train the reward model, while the remaining 15k data were used to train PPO. The results indicate that both previous RL and SFT methods exhibit limited effectiveness in directly fitting the final response, with some benchmarks even showing performance degradation (e.g. DocVQA, DUDE, etc.). In contrast, our training approach shown in Tab.~\ref{table:mllm}, which incorporates the CoT reasoning in RL method, consistently improves model performance across all benchmarks.

\noindent\textbf{Data Generation Pipeline.}
Our data generation process follows a self-evolution paradigm and adopts an actor-critics framework, where the actor is the SFTed model and the critics refer to the pre-SFT model. In this study, we conduct ablation experiments on several modules within the pipeline. As shown in Tab.~\ref{table:data}, our objective is to maximize the proportion of positive instances in the constructed preference dataset, specifically WP-LN (win-positive, lose-negative), while minimizing the proportion of negative instances, WN-LP (win-negative, lose-positive). Experimental results indicate that under our pipeline, the proportion of positive instances is \textbf{64.64\%}. After the first and second iterations -- where the model undergoes full \method training and then participates in data construction again -- this proportion increases by \textbf{3.18\%} and \textbf{2.30\%}, respectively. Besides, the number of valid data increases from 30k to 33k to \textbf{35k}. This is because, through iterative training, the model's capability improves to overcome bottlenecks where obtaining the correct answer was previously outside of its ability. Specifically, \emph{some questions are inherently too difficult for model previously, meaning that no matter how many times the model responds, the answer remains incorrect}, making it impossible to construct win/lose preference.

When replacing the critics with GPT-4o, the proportion of positive instances remains largely unchanged (64.64\% vs. 65.31\%), but this comes at the cost of increased token consumption and reduced response speed. Conversely, when substituting the critics with the SFTed model itself, the data volume decreases to \textbf{1/10} of the original (3k vs. 30k), accompanied by a \textbf{9.96\%} reduction in the proportion of positive instances. Furthermore, omitting the evaluation of the generated bounding boxes results in a \textbf{33.62\%} drop in the proportion of positive instances, while the absence of the diversity controller leads to a \textbf{12.62\%} reduction. Considering these factors, we adopt the approach described in Sec.~\ref{sec: data-generation}.

\input{Tabs/tab_data}

%% file: Tabs/tab_baseline.tex
\begin{table*}[t!]
    \caption{The evaluation of different baselines on MME~\cite{mme}, MMBench~\cite{mmb}, and POPE~\cite{pope} datasets. Datasets marked with [D] are dense-labeled datasets (e.g., CoT data). In different methods, [B] denotes the base model, [D] represents data-driven SFT methods, and [T] refers to tool-usage methods (e.g., agents). The \textbf{best} is highlighted and the second-best is \underline{underlined}. Remark: the data number considered here includes only the data used to enhance specific model capabilities and pretraining data, excluding general instruction-tuning dataset. }\label{table:baseline}
    \vspace{-0.8em}
    \centering
    \footnotesize
    \setlength{\tabcolsep}{2mm}
    \begin{tabular}{lcccccc}
    \hline
    \textbf{Method}          & \textbf{LLM}  & \textbf{Vision Encoder} & \textbf{MME}    & \textbf{MMBench} & \textbf{POPE} & \textbf{Dataset Num.}   \\ \hline
    LLaVA {[}B{]}~\cite{mllm:liu2023llava}            & Vicuna-7B~\cite{vicuna}     & CLIP-ViT-L-14-224~\cite{clip}       & 1051.2          & {34.4}       & 76.5          & 558K                    \\
    SEAL  {[}D{]}~\cite{rela:vstar}            & Vicuna-7B     & CLIP-ViT-L-14-224       & 1128.9          & 33.1             & \textbf{82.4} & 558K + 387K {[}D{]}     \\
    LLaVA + P2G {[}T{]}~\cite{rela:plugandplay}     & Vicuna-7B     & CLIP-ViT-L-14-224       & \underline{1223.0} & —                & —             & 558K + 427K {[}D{]}     \\
    \rowcolor[HTML]{E7E6E6} 
    LLaVA + \method               & Vicuna-7B     & CLIP-ViT-L-14-224       & {1183.8}    & \underline{37.5}    & {78.2}    & 558K + 30K {[}D{]}+180K \\
    \rowcolor[HTML]{E7E6E6} 
    LLaVA + \method -- Iter1             & Vicuna-7B     & CLIP-ViT-L-14-224       & {\textbf{1238.3}}    & \textbf{\textbf{38.6}}    & {\underline{80.4}}    & 180K \\ \hline
    LLaVA-1.5  {[}B{]}~\cite{mllm:liu2023improvedllava}       & Vicuna-7B     & CLIP-ViT-L-14-336       & {1510.7}    & 64.3             & 85.8          & 558K                    \\
    VisCoT {[}D{]}~\cite{shao2024visual}           & Vicuna-7B     & CLIP-ViT-L-14-336       & 1453.6          & {67.9}       & {86.0}    & 558K + 376K {[}D{]}     \\
    \rowcolor[HTML]{E7E6E6} 
    LLaVA-1.5 + \method           & Vicuna-7B     & CLIP-ViT-L-14-336       & \underline{1526.3} & \underline{70.1}    & \underline{87.5} & 558K + 30K {[}D{]}+180K \\
    \rowcolor[HTML]{E7E6E6} 
    LLaVA-1.5 + \method -- Iter1            & Vicuna-7B     & CLIP-ViT-L-14-336       & \textbf{1560.0} & \textbf{71.7}    & \textbf{88.8} & 180K \\ \hline
    LLaVA-NeXT  {[}B{]}~\cite{mllm:liu2024llavanext}      & Vicuna-7B-1.5~\cite{vicuna1.5} & CLIP-ViT-L-14-336       & {1611.1}    & 72.3             & —             & 558K                    \\
    VisionLLM v2  {[}D{]}~\cite{rela:wu2025visionllmv2}    & Vicuna-7B-1.5 & CLIP-ViT-L-14-336       & 1512.5          & 77.1             & {87.5}    & 892K                    \\
    Insight-V-LLaVA  {[}T{]}~\cite{rela:insight-v} & Vicuna-7B-1.5 & CLIP-ViT-L-14-336       & 1583.9          & \textbf{81.7}    & —             & 558K + 215K {[}D{]}     \\
    \rowcolor[HTML]{E7E6E6} 
    LLaVA-NeXT + \method           & Vicuna-7B-1.5 & CLIP-ViT-L-14-336       & \underline{1619.2} & {78.8}       & \underline{88.4} & 558K + 30K {[}D{]}+180K \\ 
    \rowcolor[HTML]{E7E6E6} 
    LLaVA-NeXT + \method -- Iter1           & Vicuna-7B-1.5 & CLIP-ViT-L-14-336       & \textbf{1637.0} & {\underline{80.0}}       & \textbf{89.3} & 180K \\ \hline
    \end{tabular}
    \vspace{-0.8em}
\end{table*}

%% file: Tabs/tab_mllm.tex
\begin{table*}[t!]
    \caption{Performance on the different benchmarks. The amount of dense-labeled CoT data with bounding box annotations used is indicated in []. The \textbf{best} results from different LMMs are highlighted.}\label{table:mllm}
    \vspace{-0.8em}
    \centering
    \scriptsize
    \setlength{\tabcolsep}{0.7mm}
    \begin{tabular}{lccccccccccccc}
    \hline
                         & \multicolumn{1}{l}{}                  & \multicolumn{5}{c}{\textbf{Doc/Text}}                                                                                                                                                                      & \textbf{Chart}                         & \textbf{General VQA}                   & \multicolumn{3}{c}{\textbf{Relation Reasoning}}                                                                          & \textbf{Fine-grained}                  & \multicolumn{1}{l}{}                   \\
    \textbf{LMM}        & \textbf{Training Phase}               & DocVQA                                 & TextCaps                               & TextVQA                                & DUDE                                   & SROIE                                  & InfogVQA                        & Flickr30k                              & GQA                                    & Open images                            & VSR                                    & CUB                         & \textbf{Avg}                           \\ \hline
                         & Base (w/o CoT)                        & 0.244                                  & 0.597                                  & 0.588                                  & 0.290                                  & 0.136                                  & 0.400                                  & 0.581                                  & 0.534                                  & 0.412                                  & 0.572                                  & 0.530                                  & 0.444                                  \\
                         & \cellcolor[HTML]{E9F7EF}VisCoT [438k]~\cite{shao2024visual} & \cellcolor[HTML]{E9F7EF}0.355          & \cellcolor[HTML]{E9F7EF}0.610          & \cellcolor[HTML]{E9F7EF}0.719          & \cellcolor[HTML]{E9F7EF}0.279          & \cellcolor[HTML]{E9F7EF}0.341          & \cellcolor[HTML]{E9F7EF}0.356          & \cellcolor[HTML]{E9F7EF}0.671          & \cellcolor[HTML]{E9F7EF}0.616          & \cellcolor[HTML]{E9F7EF}0.833          & \cellcolor[HTML]{E9F7EF}0.682          & \cellcolor[HTML]{E9F7EF}0.556          & \cellcolor[HTML]{E9F7EF}0.547          \\
    LLaVA-1.5-7B~\cite{mllm:liu2023improvedllava}         & SFT [30k]                                   & 0.336                                  & 0.597                                  & 0.715                                  & 0.270                                  & 0.308                                  & 0.336                                  & 0.671                                  & 0.617                                  & 0.833                                  & 0.676                                  & 0.559                                  & 0.538                                  \\
                         & SFT+RL1                               & 0.382                                  & 0.612                                  & 0.724                                  & 0.300                                  & 0.378                                  & 0.406                                  & 0.674                                  & 0.639                                  & 0.838                                  & 0.715                                  & 0.579                                  & 0.568                                  \\
                         & \cellcolor[HTML]{E7E6E6}SFT+RL1+RL2   & \cellcolor[HTML]{E7E6E6}\textbf{0.419} & \cellcolor[HTML]{E7E6E6}\textbf{0.641} & \cellcolor[HTML]{E7E6E6}\textbf{0.759} & \cellcolor[HTML]{E7E6E6}\textbf{0.394} & \cellcolor[HTML]{E7E6E6}\textbf{0.411} & \cellcolor[HTML]{E7E6E6}\textbf{0.497} & \cellcolor[HTML]{E7E6E6}\textbf{0.675} & \cellcolor[HTML]{E7E6E6}\textbf{0.666} & \cellcolor[HTML]{E7E6E6}\textbf{0.848} & \cellcolor[HTML]{E7E6E6}\textbf{0.748} & \cellcolor[HTML]{E7E6E6}\textbf{0.598} & \cellcolor[HTML]{E7E6E6}\textbf{0.605} \\
                         & Base (w/o CoT)                        & 0.431                                  & 0.586                                  & 0.570                                  & 0.332                                  & 0.114                                  & 0.361                                  & 0.525                                  & 0.559                                  & 0.462                                  & 0.594                                  & 0.520                                  & 0.459                                  \\
    LLaVA-NeXT-7B~\cite{mllm:liu2024llavanext}        & SFT [30k]                                   & 0.423                                  & 0.580                                  & 0.722                                  & 0.330                                  & 0.293                                  & 0.356                                  & 0.589                                  & 0.684                                  & 0.821                                  & 0.767                                  & 0.551                                  & 0.556                                  \\
                         & SFT+RL1                               & 0.474                                  & 0.611                                  & 0.728                                  & 0.373                                  & 0.350                                  & 0.447                                  & \textbf{0.592}                         & 0.707                                  & 0.826                                  & 0.837                                  & 0.573                                  & 0.593                                  \\
                         & \cellcolor[HTML]{E7E6E6}SFT+RL1+RL2   & \cellcolor[HTML]{E7E6E6}\textbf{0.508} & \cellcolor[HTML]{E7E6E6}\textbf{0.655} & \cellcolor[HTML]{E7E6E6}\textbf{0.743} & \cellcolor[HTML]{E7E6E6}\textbf{0.474} & \cellcolor[HTML]{E7E6E6}\textbf{0.379} & \cellcolor[HTML]{E7E6E6}\textbf{0.525} & \cellcolor[HTML]{E7E6E6}\textbf{0.592} & \cellcolor[HTML]{E7E6E6}\textbf{0.738} & \cellcolor[HTML]{E7E6E6}\textbf{0.837} & \cellcolor[HTML]{E7E6E6}\textbf{0.871} & \cellcolor[HTML]{E7E6E6}\textbf{0.587} & \cellcolor[HTML]{E7E6E6}\textbf{0.628} \\
                         & Base (w/o CoT)                        & 0.797                                  & 0.771                                  & 0.879                                  & 0.588                                  & 0.629                                  & 0.637                                  & 0.601                                  & 0.484                                  & 0.335                                  & 0.589                                  & 0.674                                  & 0.635                                  \\
    Llama-3.2-V-11B~\cite{llama3.2} & SFT [30k]                                   & 0.776                                  & 0.762                                  & 0.880                                  & 0.584                                  & 0.634                                  & 0.633                                  & 0.712                                  & 0.683                                  & 0.728                                  & 0.720                                  & 0.855                                  & 0.724                                  \\
                         & SFT+RL1                               & 0.811                                  & 0.791                                  & 0.890                                  & 0.599                                  & 0.698                                  & 0.688                                  & 0.724                                  & 0.707                                  & 0.731                                  & 0.738                                  & 0.864                                  & 0.749                                  \\
                         & \cellcolor[HTML]{E7E6E6}SFT+RL1+RL2   & \cellcolor[HTML]{E7E6E6}\textbf{0.844} & \cellcolor[HTML]{E7E6E6}\textbf{0.835} & \cellcolor[HTML]{E7E6E6}\textbf{0.897} & \cellcolor[HTML]{E7E6E6}\textbf{0.638} & \cellcolor[HTML]{E7E6E6}\textbf{0.733} & \cellcolor[HTML]{E7E6E6}\textbf{0.714} & \cellcolor[HTML]{E7E6E6}\textbf{0.731} & \cellcolor[HTML]{E7E6E6}\textbf{0.757} & \cellcolor[HTML]{E7E6E6}\textbf{0.794} & \cellcolor[HTML]{E7E6E6}\textbf{0.822} & \cellcolor[HTML]{E7E6E6}\textbf{0.884} & \cellcolor[HTML]{E7E6E6}\textbf{0.786} \\
                         & Base (w/o CoT)                        & 0.528                                  & 0.504                                  & 0.548                                  & 0.125                                  & 0.114                                  & 0.220                                  & 0.534                                  & 0.561                                  & 0.462                                  & 0.585                                  & 0.529                                  & 0.428                                  \\
    MiniCPM-o-2.6-8B~\cite{yao2024minicpm}     & SFT [30k]                                   & 0.518                                  & 0.498                                  & 0.551                                  & 0.134                                  & 0.133                                  & 0.239                                  & 0.615                                  & 0.727                                  & 0.789                                  & 0.787                                  & 0.715                                  & 0.519                                  \\
                         & SFT+RL1                               & 0.551                                  & 0.533                                  & 0.561                                  & 0.150                                  & 0.182                                  & 0.286                                  & 0.630                                  & 0.737                                  & 0.799                                  & 0.824                                  & 0.734                                  & 0.544                                  \\
                         & \cellcolor[HTML]{E7E6E6}SFT+RL1+RL2   & \cellcolor[HTML]{E7E6E6}\textbf{0.596} & \cellcolor[HTML]{E7E6E6}\textbf{0.600} & \cellcolor[HTML]{E7E6E6}\textbf{0.565} & \cellcolor[HTML]{E7E6E6}\textbf{0.209} & \cellcolor[HTML]{E7E6E6}\textbf{0.251} & \cellcolor[HTML]{E7E6E6}\textbf{0.353} & \cellcolor[HTML]{E7E6E6}\textbf{0.639} & \cellcolor[HTML]{E7E6E6}\textbf{0.793} & \cellcolor[HTML]{E7E6E6}\textbf{0.870} & \cellcolor[HTML]{E7E6E6}\textbf{0.864} & \cellcolor[HTML]{E7E6E6}\textbf{0.756} & \cellcolor[HTML]{E7E6E6}\textbf{0.591} \\
                         & Base (w/o CoT)                        & 0.017                                  & 0.498                                  & 0.536                                  & 0.129                                  & 0.114                                  & 0.197                                  & 0.529                                  & 0.558                                  & 0.486                                  & 0.543                                  & 0.541                                  & 0.377                                  \\
    PaliGemma2-10B~\cite{steiner2024paligemma}       & SFT [30k]                                   & 0.110                                  & 0.498                                  & 0.544                                  & 0.134                                  & 0.133                                  & 0.225                                  & 0.611                                  & 0.718                                  & 0.800                                  & 0.770                                  & 0.724                                  & 0.479                                  \\
                         & SFT+RL1                               & 0.169                                  & 0.527                                  & 0.549                                  & 0.163                                  & 0.179                                  & 0.272                                  & 0.621                                  & 0.731                                  & 0.811                                  & 0.822                                  & 0.736                                  & 0.507                                  \\
                         & \cellcolor[HTML]{E7E6E6}SFT+RL1+RL2   & \cellcolor[HTML]{E7E6E6}\textbf{0.303} & \cellcolor[HTML]{E7E6E6}\textbf{0.585} & \cellcolor[HTML]{E7E6E6}\textbf{0.560} & \cellcolor[HTML]{E7E6E6}\textbf{0.229} & \cellcolor[HTML]{E7E6E6}\textbf{0.248} & \cellcolor[HTML]{E7E6E6}\textbf{0.336} & \cellcolor[HTML]{E7E6E6}\textbf{0.639} & \cellcolor[HTML]{E7E6E6}\textbf{0.789} & \cellcolor[HTML]{E7E6E6}\textbf{0.884} & \cellcolor[HTML]{E7E6E6}\textbf{0.847} & \cellcolor[HTML]{E7E6E6}\textbf{0.764} & \cellcolor[HTML]{E7E6E6}\textbf{0.562} \\
                         & Base (w/o CoT)                        & 0.115                                  & 0.522                                  & 0.551                                  & 0.130                                  & 0.122                                  & 0.205                                  & 0.522                                  & 0.561                                  & 0.468                                  & 0.587                                  & 0.497                                  & 0.389                                  \\
    Yi-VL-6B~\cite{llm:young2024yi}             & SFT [30k]                                   & 0.168                                  & 0.521                                  & 0.598                                  & 0.139                                  & 0.152                                  & 0.247                                  & 0.606                                  & 0.721                                  & 0.772                                  & 0.792                                  & 0.695                                  & 0.492                                  \\
                         & SFT+RL1                               & 0.208                                  & 0.564                                  & 0.610                                  & 0.174                                  & 0.182                                  & 0.294                                  & 0.613                                  & 0.747                                  & 0.799                                  & 0.844                                  & 0.713                                  & 0.523                                  \\
                         & \cellcolor[HTML]{E7E6E6}SFT+RL1+RL2   & \cellcolor[HTML]{E7E6E6}\textbf{0.318} & \cellcolor[HTML]{E7E6E6}\textbf{0.611} & \cellcolor[HTML]{E7E6E6}\textbf{0.627} & \cellcolor[HTML]{E7E6E6}\textbf{0.234} & \cellcolor[HTML]{E7E6E6}\textbf{0.280} & \cellcolor[HTML]{E7E6E6}\textbf{0.358} & \cellcolor[HTML]{E7E6E6}\textbf{0.620} & \cellcolor[HTML]{E7E6E6}\textbf{0.804} & \cellcolor[HTML]{E7E6E6}\textbf{0.853} & \cellcolor[HTML]{E7E6E6}\textbf{0.871} & \cellcolor[HTML]{E7E6E6}\textbf{0.726} & \cellcolor[HTML]{E7E6E6}\textbf{0.573} \\
                         & Base (w/o CoT)                        & 0.836                                  & 0.760                                  & 0.847                                  & 0.606                                  & 0.789                                  & 0.685                                  & 0.601                                  & 0.467                                  & 0.289                                  & 0.581                                  & 0.583                                  & 0.640                                  \\
    Qwen2.5-VL-7B~\cite{bai2025qwen2.5}        & SFT [30k]                                   & 0.807                                  & 0.720                                  & 0.886                                  & 0.580                                  & 0.719                                  & 0.635                                  & 0.630                                  & 0.626                                  & 0.764                                  & 0.782                                  & 0.876                                  & 0.730                                  \\
                         & SFT+RL1                               & 0.842                                  & 0.768                                  & 0.895                                  & 0.600                                  & 0.784                                  & 0.692                                  & 0.642                                  & 0.669                                  & 0.788                                  & 0.822                                  & 0.888                                  & 0.763                                  \\
                         & \cellcolor[HTML]{E7E6E6}SFT+RL1+RL2   & \cellcolor[HTML]{E7E6E6}\textbf{0.874} & \cellcolor[HTML]{E7E6E6}\textbf{0.819} & \cellcolor[HTML]{E7E6E6}\textbf{0.897} & \cellcolor[HTML]{E7E6E6}\textbf{0.640} & \cellcolor[HTML]{E7E6E6}\textbf{0.829} & \cellcolor[HTML]{E7E6E6}\textbf{0.753} & \cellcolor[HTML]{E7E6E6}\textbf{0.675} & \cellcolor[HTML]{E7E6E6}\textbf{0.700} & \cellcolor[HTML]{E7E6E6}\textbf{0.814} & \cellcolor[HTML]{E7E6E6}\textbf{0.864} & \cellcolor[HTML]{E7E6E6}\textbf{0.892} & \cellcolor[HTML]{E7E6E6}\textbf{0.796} \\ \hline
    \end{tabular}
    \vspace{-2em}
\end{table*}

%% file: Tabs/tab_sft_bb.tex
\begin{table}[t!]
    \caption{Ratio of successful bounding box outputs of different SFT data number in terms of Qwen2.5-VL-7B. We evaluate on Visual CoT benchmark with 8281 data.}\label{table:sft}
    \vspace{-0.8em}
    \centering
    \footnotesize
    \setlength{\tabcolsep}{1.7mm}
    \begin{tabular}{cccccc}
    \hline
    SFT Data Num.         & 10k                  & 20k                  & 30k                  & 50k                  & 100k                 \\ \hline
    Ratio                & 28.14\%              & 95.48\%              & 99.87\%              & 99.87\%              & 99.87\%              \\ \hline
    \multicolumn{1}{l}{} & \multicolumn{1}{l}{} & \multicolumn{1}{l}{} & \multicolumn{1}{l}{} & \multicolumn{1}{l}{} & \multicolumn{1}{l}{}
    \end{tabular}
    \vspace{-2em}
\end{table}

%% file: Tabs/tab_detection.tex
\begin{table*}[t!]
    \caption{Detection performance (Top-1 Accuracy@0.5) on the various benchmark, where both "Ours" and "Visual-CoT" similarly utilize LLaVA-1.5-7B as the base model. The amount of dense-labeled data with bounding box annotations is indicated in []. The \textbf{best} is highlighted.}\label{table:detection}
    \vspace{-0.8em}
    \centering
    \footnotesize
    \setlength{\tabcolsep}{0.8mm}
    \begin{tabular}{ccccccccccccc}
    \hline
    \multicolumn{1}{l}{} & \multicolumn{5}{c}{\textbf{Doc/Text}}                                              & \textbf{Chart} & \textbf{General VQA} & \multicolumn{3}{c}{\textbf{Relation Reasoning}}  & \textbf{Fine-grained} & \multicolumn{1}{l}{} \\
    \textbf{Method}      & DocVQA         & TextCaps       & TextVQA        & DUDE           & SROIE          & InfogVQA       & Flickr30k            & GQA            & Open images    & VSR            & CUB                   & \textbf{Avg}         \\ \hline
    VisCoT [438k]~\cite{shao2024visual}     & 0.136          & 0.413          & 0.468          & 0.050          & 0.157          & 0.072          & 0.496                & 0.420          & 0.576          & 0.696          & 0.670                 & 0.378                \\
    \method-SFT [30k]            & 0.133          & 0.406          & 0.460          & 0.048          & 0.146          & 0.069          & 0.495                & 0.421          & 0.575          & 0.691          & 0.673                 & 0.374                \\
    \method-SFT+RL1         & 0.190          & 0.433          & 0.471          & 0.081          & 0.216          & 0.133          & 0.518                & 0.450          & 0.581          & 0.733          & 0.701                 & 0.410                \\
    \rowcolor[HTML]{E7E6E6} 
    \method-SFT+RL1+RL2     & \textbf{0.231} & \textbf{0.464} & \textbf{0.504} & \textbf{0.124} & \textbf{0.252} & \textbf{0.172} & \textbf{0.519}       & \textbf{0.487} & \textbf{0.590} & \textbf{0.745} & \textbf{0.715}        & \textbf{0.437}       \\ \hline
    \end{tabular}
    \vspace{-0.8em}
\end{table*}

%% file: Tabs/tab_strategy.tex
\begin{table*}
    \caption{Different training strategies for directly fitting the final response in terms of Qwen2.5-VL-7B. Besides, we also ablated on RL2-Only. The red background indicates a decline in performance compared to the original model, while the \textbf{best} is highlighted.}\label{table:strategy}
    \vspace{-0.8em}
    \centering
    \footnotesize
    \setlength{\tabcolsep}{1.1mm}
    \begin{tabular}{lcccccccccccc}
    \hline
                    & \multicolumn{5}{c}{\textbf{Doc/Text}}                                                                                                                                                                      & \textbf{Chart}                         & \textbf{General VQA}                   & \multicolumn{3}{c}{\textbf{Relation Reasoning}}                                                                          & \textbf{Fine-grained}                  & \multicolumn{1}{l}{}                   \\
    \textbf{Method} & DocVQA                                 & TextCaps                               & TextVQA                                & DUDE                                   & SROIE                                  & InfogVQA                               & Flickr30k                              & GQA                                    & Open images                            & VSR                                    & CUB                                    & \textbf{Avg}                           \\ \hline
    Ori             & 0.836                                  & 0.760                                  & 0.847                                  & 0.606                                  & 0.789                                  & 0.685                                  & 0.601                                  & 0.467                                  & 0.289                                  & 0.581                                  & 0.583                                  & 0.640                                  \\
    SFT             & \cellcolor[HTML]{FFE9E8}0.778          & \cellcolor[HTML]{FFE9E8}0.751          & 0.856                                  & \cellcolor[HTML]{FFE9E8}0.566          & \cellcolor[HTML]{FFE9E8}0.713          & \cellcolor[HTML]{FFE9E8}0.631          & 0.627                                  & 0.524                                  & 0.345                                  & 0.626                                  & 0.736                                  & 0.650                                  \\
    DPO~\cite{rafailov2024direct}             & \cellcolor[HTML]{FFE9E8}0.791          & 0.793                                  & \cellcolor[HTML]{FFE9E8}0.837          & \cellcolor[HTML]{FFE9E8}0.599          & \cellcolor[HTML]{FFE9E8}0.776          & \cellcolor[HTML]{FFE9E8}0.678          & 0.670                                  & 0.470                                  & 0.298                                  & 0.594                                  & 0.707                                  & 0.656                                  \\
    KTO~\cite{kto}             & \cellcolor[HTML]{FFE9E8}0.797          & 0.790                                  & \cellcolor[HTML]{FFE9E8}0.833          & \cellcolor[HTML]{FFE9E8}0.597          & \cellcolor[HTML]{FFE9E8}0.780          & 0.694                                  & 0.671                                  & 0.469                                  & 0.299                                  & 0.599                                  & 0.707                                  & 0.658                                  \\
    PPO~\cite{ppo}             & \cellcolor[HTML]{FFE9E8}0.794          & 0.791                                  & \cellcolor[HTML]{FFE9E8}0.837          & \cellcolor[HTML]{FFE9E8}0.597          & \cellcolor[HTML]{FFE9E8}0.787          & 0.700                                  & 0.672                                  & 0.472                                  & 0.302                                  & 0.606                                  & 0.713                                  & 0.661     \\
    \method-RL2-Only             & \cellcolor[HTML]{FFE9E8}0.816          & 0.769                                & 0.897          & \cellcolor[HTML]{FFE9E8}0.602          & 0.800          & \cellcolor[HTML]{FFE9E8}0.681                                  & 0.637                               & 0.652                                  & 0.786                                  & 0.827                                  & 0.882                                  & 0.759     \\
    \cellcolor[HTML]{E7E6E6}\method-Full       & \cellcolor[HTML]{E7E6E6}\textbf{0.874} & \cellcolor[HTML]{E7E6E6}\textbf{0.819} & \cellcolor[HTML]{E7E6E6}\textbf{0.897} & \cellcolor[HTML]{E7E6E6}\textbf{0.640} & \cellcolor[HTML]{E7E6E6}\textbf{0.829} & \cellcolor[HTML]{E7E6E6}\textbf{0.753} & \cellcolor[HTML]{E7E6E6}\textbf{0.675} & \cellcolor[HTML]{E7E6E6}\textbf{0.700} & \cellcolor[HTML]{E7E6E6}\textbf{0.814} & \cellcolor[HTML]{E7E6E6}\textbf{0.864} & \cellcolor[HTML]{E7E6E6}\textbf{0.892} & \cellcolor[HTML]{E7E6E6}\textbf{0.796} \\ \hline
    \end{tabular}
    \vspace{-2em}
\end{table*}

%% file: Figs/chart.tex
\begin{figure}[!t]
    \vspace{-0.3em}
    \centering
    \includegraphics[width=0.5\textwidth]{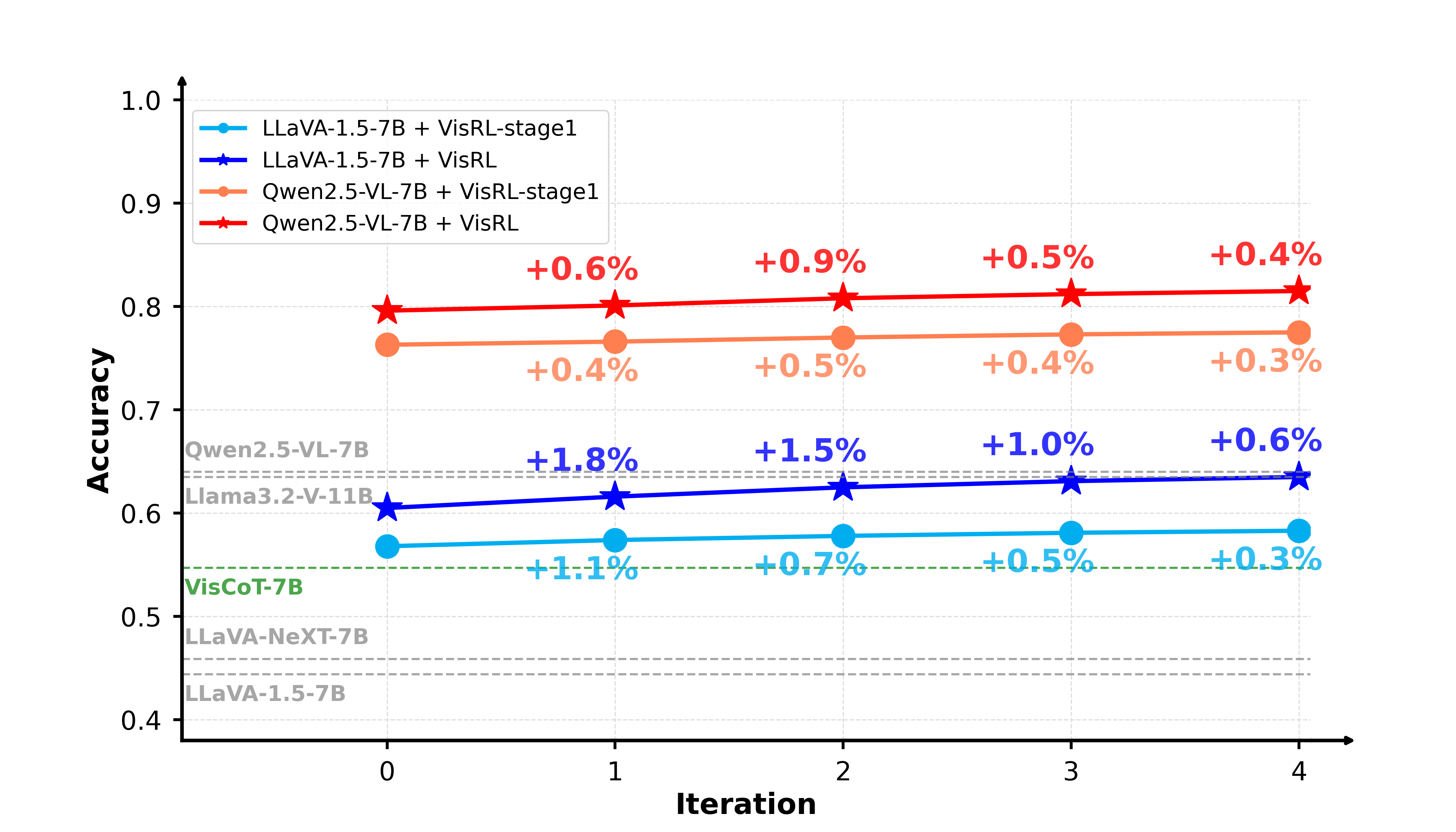}
    \caption{Performance of our VisRL over multiple iterations, attributing to the intertwined improvement of data quality and model capability during the iterative process. The accuracy is calculated as the average value over the 11 datasets listed in Tab.~\ref{table:mllm}.}\label{fig:chart}
    \vspace{-2em}
\end{figure}

%% file: Figs/comparison.tex
\begin{figure}[!htbp]
    \centering
    \includegraphics[width=0.46\textwidth]{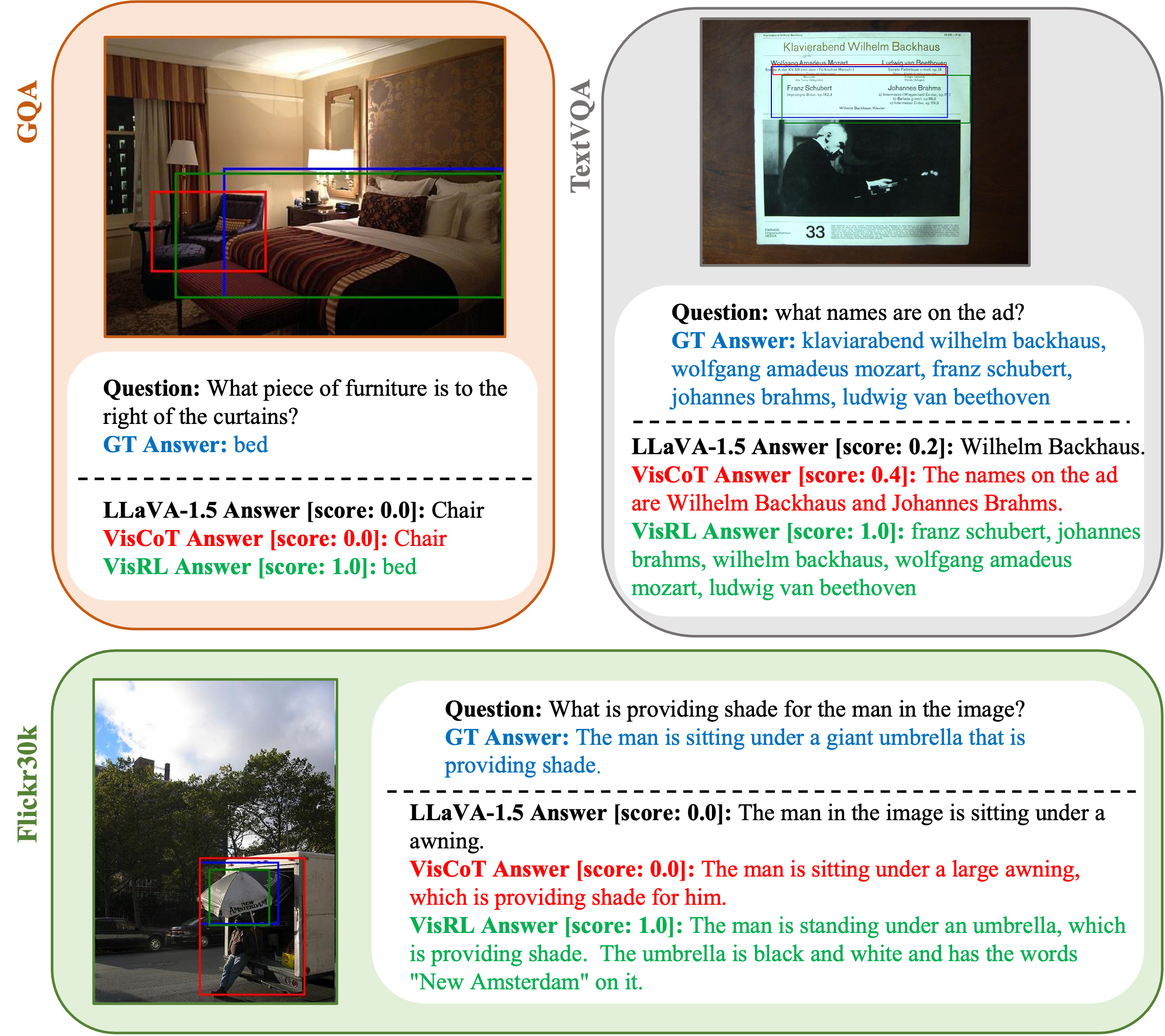}
    \vspace{-0.8em}
    \caption{Visualization of LLaVa-1.5 vs. VisCoT vs. VisRL (based on LLaVa-1.5). GT bounding boxes are shown in \textcolor{blue}{blue}, VisCoT-generated bounding boxes are shown in \textcolor{red}{red}, while Ours-generated bounding boxes are in \textcolor{green}{green}. The scores are evaluated by the GPT-4o. Our method consistently delivers the best results across various benchmarks. More visualizations are in the Supp. Mat..}\label{fig:comparison}
    \vspace{-1.6em}
\end{figure}

%% file: Tabs/tab_data.tex
\begin{table}
    \caption{Ablation of our data generation pipeline in terms of Qwen2.5-VL-7B. We evaluate the data quality by comparing the IoU (Top-1 Accuracy@0.5) between the annotated preference data's bounding boxes and the GT bounding boxes. "W" and "L" represent the win and loss in the data annotation, respectively, while "P" and "N" indicate positive or negative. Ideally, we expect WP-LN to be as large as possible, as highlighted in green.}\label{table:data}
    \vspace{-0.8em}
    \centering
    \footnotesize
    \setlength{\tabcolsep}{0.8mm}
    \begin{tabular}{lc
    >{\columncolor[HTML]{EAFAF1}}c ccc}
    \hline
                                 & WP-LP           & WP-LN            & WN-LP           & WN-LN            & Data Num. \\ \hline
    w GPT-4o-2024-11-20         & 0.00\%          & 65.31\%          & 1.32\%          & 33.37\%          & \textbf{47k}            \\
    w SFTed Model                & 0.00\%          & 54.68\%          & 0.00\%          & 45.32\%          & 3k                      \\
    \scriptsize{w/o Bounding Box Critics}     & 5.42\%          & 31.02\%          & 10.04\%         & 53.51\%          & 86k                     \\
    \scriptsize{w/o Diversity Controller} & 4.53\%          & 52.02\%          & 4.68\%          & 38.77\%          & 19k                     \\
    \method-Full                    & 0.43\%          & 64.64\%          & 1.64\%          & 33.29\%          & 30k                     \\
    \method-Full-Iter1              & 0.45\%          & 67.82\%          & 0.82\%          & 30.91\%          & 33k                     \\
    \method-Full-Iter2              & \textbf{0.47\%} & \textbf{70.12\%} & \textbf{0.00\%} & \textbf{29.41\%} & 35k                     \\ \hline
    \end{tabular}
    \vspace{-2.5em}
\end{table}

%% file: Sec/conclusion.tex
\section{Conclusion}\label{sec:conclusion}
In this paper, we propose \method, a framework for learning intention-driven visual perception abilities from task feedback. This approach enables the model to undergo RL through self-evolution when trained on simple data with only final responses. Specifically, we introduce a novel pipeline for generating CoT preference data based on the model’s own actor-critic process, eliminating the need for external models or human annotations. Using these self-constructed data, we further optimize visual perception in two RL stages: (1) independently optimizing generated bounding box; (2) jointly optimizing both generated bounding box and response. Extensive experiments on various benchmarks and LMMs demonstrate the effectiveness of the proposed framework, establishing a solid foundation for future exploration.

%% file: Sec/supmat.tex
\label{sec:appendix}
In this supplementary material, we provide more technical details and experimental results, including 1) Detailed descriptions of dataset used in Sec.~\ref{sec:dataset} and Tab.~\ref{table:data_num}; 2) Visual grounding ability tested on REC benchmarks in Sec.~\ref{sec:grounding} and Tab.~\ref{table:coco}; 3) Our prompr designed for critics of data generation pipeline in Sec.~\ref{sec:instruction}; as well as 4): More visualization of different datasets from VisualCoT benchmarks in Sec.~\ref{sec:visualization}.

\section{Dataset}\label{sec:dataset}
\subsection{VisCoT Dataset}
We utilize the data from VisCoT~\cite{shao2024visual} and follow its predefined training/testing split. Specifically, a subset of the training set is selected for training our VisRL model, as shown in Tab.~\ref{table:data_num}. Besides, the test set remains consistent with VisCoT, as presented in Tab.~\ref{table:baseline}, Tab.~\ref{table:mllm}, Tab.~\ref{table:detection} in the main text, etc..

\noindent\textbf{Text/Doc:} 
There are five text-related datasets—TextVQA~\cite{textvqa}, DocVQA~\cite{docvqa}, DUDE~\cite{dude}, TextCaps~\cite{textcaps}, and SROIE~\cite{sroie}, covering text recognition and comprehension in various images and documents. 

\noindent\textbf{Fine-Grained Understanding:} 
The Birds-200-2011 dataset (CUB)~\cite{cub} is a widely used benchmark for fine-grained visual categorization. It includes rich visual data, detailed annotations of bird parts and attributes, and bounding boxes. To leverage this better for LMM,~\cite{shao2024visual} design questions that challenge the model to identify specific bird characteristics, testing its ability to recognize fine-grained details.

\noindent\textbf{General VQA:} 
Flickr30k~\cite{flickr30k} and Visual7W~\cite{visual7w} are used for general VQA tasks. Specifically, Flickr30k provides five captions per image and bounding boxes for most mentioned objects.~\cite{shao2024visual} further use GPT-4 to generate questions focusing on small objects, while Visual7W has already included question-answer pairs with object-level grounding annotations.

\noindent\textbf{Charts:} InfographicsVQA~\cite{infographicvqa} dataset features high-resolution infographics, to train LMMs in locating answers precisely.

\noindent\textbf{Relation Reasoning:} The Visual Spatial Reasoning (VSR)~\cite{vsr}, GQA ~\cite{gqa}, and Open Images~\cite{openimages} datasets, which are rich in spatial relational information among image objects, are used for relation-reasoning tasks.

\subsection{Comprehensive Benchmarks}
We conducted evaluations on three further benchmarks as shown in Tab.~\ref{table:baseline} in the main text: MME~\cite{mme}, which comprehensively assesses perception and cognitive abilities across 14 sub-tasks; MMBench~\cite{mmb}, a systematically designed objective benchmark for the robust and holistic evaluation, covering 20 capability dimensions; and POPE~\cite{pope}, which reframes hallucination evaluation as a series of binary questions requiring the model to determine the presence of objects in an image.

\input{Tabs/tab_data_num}
\input{Tabs/tab_coco}
\section{Visual Grounding}\label{sec:grounding}
Furthermore, we conducted additional evaluations of our VisRL on REC benchmarks. Specifically, we tested different methods on RefCOCO~\cite{refcoco} and RefCOCO+~\cite{refcoco+/g}, both of which were collected in an interactive gaming interface and follow the validation/test-A/test-B split. In these two datasets, test-A always consists of images containing multiple people, whereas test-B includes all other objects. Additionally, compared to RefCOCO, queries in RefCOCO+ do not contain absolute spatial terms, such as references to an object's location within the image (e.g., "on the right side"). RefCOCOg~\cite{refcoco+/g} was another dataset collected in a non-interactive setting, and its queries are generally longer than those in RefCOCO and RefCOCO+.

As shown in Tab.~\ref{table:coco}, VisRL surpasses all previous generalist models, even outperforming models with significantly larger parameters. Moreover, in most of cases, our method exceeds the performance of previous state-of-the-art specialist models, e.g. G-DINO-L~\cite{groundingdino} and UNINEXT~\cite{uninext}. This demonstrates the exceptional capability of our approach in accurately predicting bounding boxes. Notably, our model achieves improvements of \textbf{1\%} to \textbf{5\%} to VisCoT. "Top-1 Accuracy@0.5," refers to the accuracy of a model in correctly predicting the bounding box as the top-ranked output when the IoU between the predicted and GT bounding boxes is at least 50\%.

\section{Instruction for Critics}\label{sec:instruction}
\subsection{Evaluation of Generated Bounding Box}
Fig.~\ref{fig:bb_judger} illustrates how we design the instruction to evaluate the bounding boxes generated by $\mathcal{M}_{SFT}$ based on $\mathcal{M}_{org}$. Specifically, given the VQA data $(Q,I,R_{GT})$, $\mathcal{M}_{SFT}$ first outputs the bounding box based on $Q$ and $I$, which is then used to crop the sub-images $I^s$. Subsequently, we assess the correlation between the generated bounding box and the GT response by prompting $(Q,R_{GT}, I^s)$ to $\mathcal{M}_{org}$ (shown in Fig.~\ref{fig:bb_judger}). Thus, we achieve the evaluated score of bounding box solely based on the GT response, without the need for extra bounding box annotations.
\input{Figs/bb_judger}
\subsection{Evaluation of Generated Response}
Fig.~\ref{fig:res_judger} presents the evaluation of responses along the sampled paths. Specifically, we prompt the model $\mathcal{M}_{org}$ to assess the generated response with GT response based on the given question/image, and assigning the score accordingly.
\input{Figs/res_judger}

\section{More visualization}\label{sec:visualization}
\input{Figs/demo1}
\input{Figs/demo2}
\input{Figs/demo3}
\input{Figs/demo4}
In Fig.~\ref{fig:demo1},~\ref{fig:demo2},~\ref{fig:demo3} and ~\ref{fig:demo4}, we provide more visualization results of our VisRL compared with VisCoT, while using the same base model -- LLaVA-1.5-7B.

%% file: Tabs/tab_data_num.tex
\begin{table*}[t!]
    \caption{We detail the number of samples used on each dataset during the SFT and RL training stages in terms of Qwen2.5-VL-7B. Specifically, SFT is trained on data with bounding box labels, while RL utilizes only the image-question-answer pairs \emph{without any additional annotations}. After our preference dataset construction, the RL data is distilled from 180k to 30k samples. Moreover, the datasets used for SFT and RL are independent (no overlap), while RL1 and RL2 share the same training dataset.}\label{table:data_num}
    \vspace{-0.8em}
    \centering
    \footnotesize
    \setlength{\tabcolsep}{1.8mm}
    \begin{tabular}{llccc}
    \hline
    \textbf{Dataset} & \textbf{Category}          & \multicolumn{1}{l}{\textbf{SFT (w. bounding box CoT)}} & \textbf{RL (w/o bounding box label)} & \textbf{RL (after data generation)} \\ \hline
    Flickr30k~\cite{flickr30k}        & General VQA                & 4000                                                   & 32500                                & 4626                                       \\
    GQA~\cite{gqa}              & Relation Reasoning         & 4000                                                   & 30000                                & 5173                                       \\
    InfographicsVQA~\cite{infographicvqa}  & Charts                     & 4055                                                   & 11000                                & 2358                                       \\
    Open Images~\cite{openimages}       & Relation Reasoning         & 3053                                                   & 40000                                & 5019                                       \\
    TextCaps~\cite{textcaps}          & Text/Doc                   & 4152                                                   & 28000                                & 4869                                       \\
    TextVQA~\cite{textvqa}          & Text/Doc                   & 3524                                                   & 15000                                & 2458                                       \\
    VSR~\cite{vsr}              & Relation Reasoning         & 1876                                                   & 1500                                 & 998                                        \\
    CUB~\cite{cub}              & Fine-Grained Understanding & 1987                                                   & 2000                                 & 1278                                       \\
    Visual7W~\cite{visual7w}              & General VQA                & 4000                                                   & 20000                                & 3267                                       \\
    \textbf{Total}   &                            & 30647                                                  & 180000                               & 30046                                      \\ \hline
    \end{tabular}
\end{table*}

%% file: Tabs/tab_coco.tex
\begin{table*}[t!]
    \caption{Performance (Top-1 Accuracy@0.5) on Referring Expression Comprehension (REC) tasks. [S] refers to specialist models, while [G] refers to generalist models. The best is \textbf{highlighted}, while the second-best is \underline{underlined}.}\label{table:coco}
    \vspace{-0.8em}
    \centering
    \setlength{\tabcolsep}{1.8mm}
    \begin{tabular}{lccccccccc}
    \hline
                                      &                                 & \multicolumn{3}{c}{\textbf{RefCOCO~\cite{refcoco}}}               & \multicolumn{3}{c}{\textbf{RefCOCO+}~\cite{refcoco+/g}}              & \multicolumn{2}{c}{\textbf{RefCOCOg}~\cite{refcoco+/g}} \\
    \multirow{-2}{*}{\textbf{Method}} & \multirow{-2}{*}{\textbf{Res.}} & \textbf{val}   & \textbf{test-A} & \textbf{test-B} & \textbf{val}   & \textbf{test-A} & \textbf{test-B} & \textbf{val-u}    & \textbf{test-u}   \\ \hline
    UNINEXT {[}S{]}~\cite{uninext}                   & $640^2$                             & {\underline{ 92.64}}    & {\underline{ 94.33}}     & \textbf{91.46}  & 85.24          & 89.63           & 79.79           & {\underline{ 88.73}}       & \textbf{89.37}    \\
    G-DINO-L {[}S{]}~\cite{groundingdino}                  & $384^2$                             & 90.56          & 93.19           & 88.24           & 82.75          & 88.95           & 75.92           & 86.13             & 87.02             \\
    OFA-L {[}G{]}~\cite{ofa}                     & $480^2$                             & 79.96          & 83.67           & 76.39           & 68.29          & 76.00           & 61.75           & 67.57             & 67.58             \\
    Shikra 7B {[}G{]}~\cite{shikra}                 & $224^2$                             & 87.01          & 90.61           & 80.24           & 81.60          & 87.36           & 72.12           & 82.27             & 82.19             \\
    MiniGPT-v2-7B {[}G{]}~\cite{minigpt}             & $448^2$                             & 88.69          & 91.65           & 85.33           & 79.97          & 85.12           & 74.45           & 84.44             & 84.66             \\
    Qwen-VL-7B {[}G{]}~\cite{llm5:bai2023qwen}                & $448^2$                             & 89.36          & 92.26           & 85.34           & 83.12          & 88.25           & 77.21           & 85.58             & 85.48             \\
    Ferret-7B {[}G{]}~\cite{ferret}                 & $336^2$                             & 87.49          & 91.35           & 82.45           & 80.78          & 87.38           & 73.14           & 83.93             & 84.76             \\
    u-LLaVA-7B {[}G{]}~\cite{u-llava}                & $224^2$                             & 80.41          & 82.73           & 77.82           & 72.21          & 76.61           & 66.79           & 74.77             & 75.63             \\
    SPHINX-13B {[}G{]}~\cite{sphinx}                & $224^2$                             & 89.15          & 91.37           & 85.13           & 82.77          & 87.29           & 76.85           & 84.87             & 83.65             \\
    VisCoT-7B~\cite{shao2024visual}                         & $336^2$                             & 91.77          & 94.25           & 87.46           & {\underline{ 87.46}}    & {\underline{ 92.05}}     & {\underline{ 81.18}}     & 88.38             & 88.34             \\
    \rowcolor[HTML]{E7E6E6} 
    LLaVA-1.5-7B~\cite{mllm:liu2023improvedllava}  + \method             & $336^2$                             & \textbf{92.72} & \textbf{96.18}  & {\underline{ 90.21}}     & \textbf{90.23} & \textbf{94.10}  & \textbf{85.77}  & \textbf{91.17}    & {\underline{ 89.28}}       \\ \hline
    \end{tabular}
\end{table*}

%% file: Figs/bb_judger.tex
\begin{figure}[!htbp]
    \centering
    \includegraphics[width=0.45\textwidth]{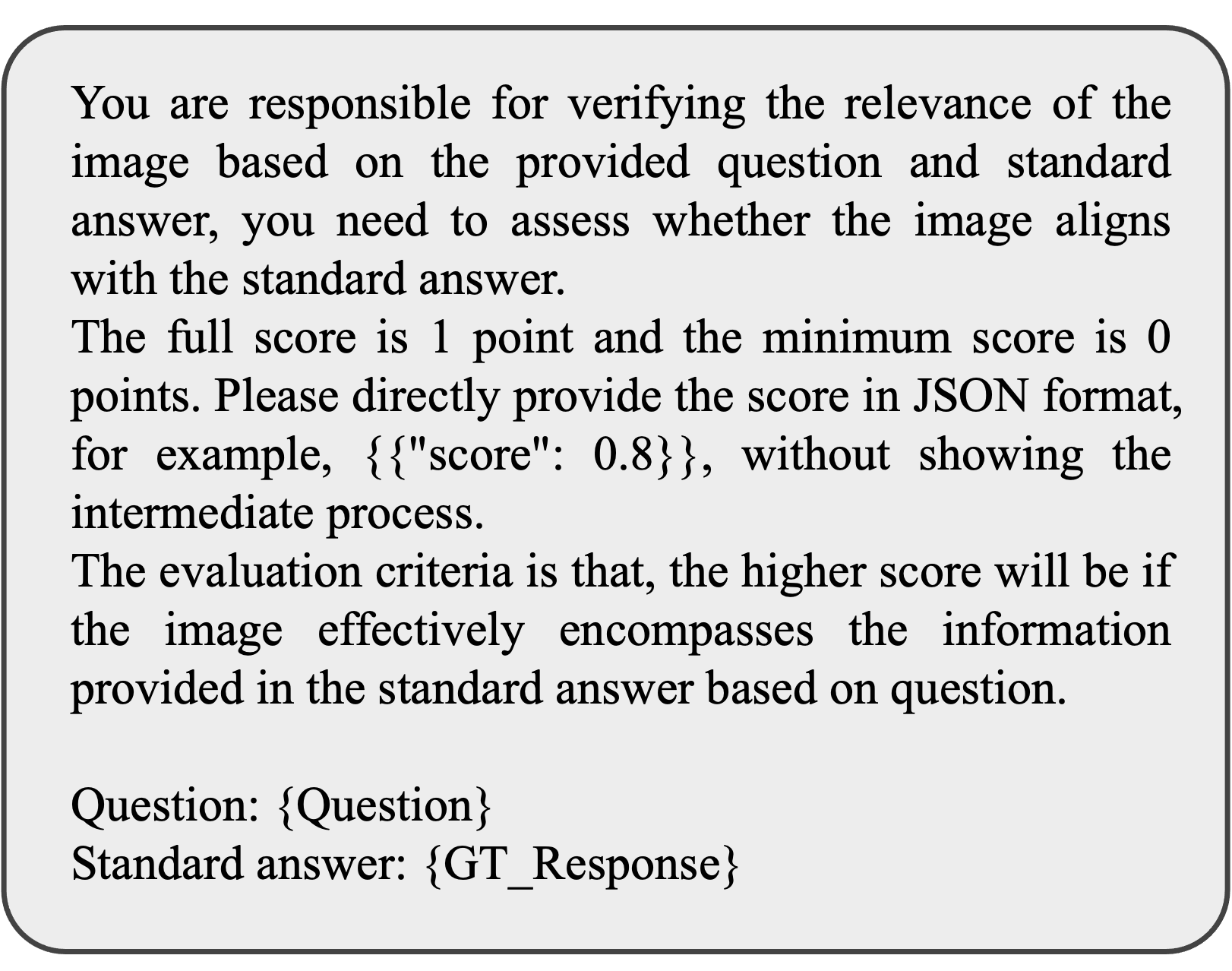}
    \caption{Prompt for the bounding box critics.}\label{fig:bb_judger}
\end{figure}

%% file: Figs/res_judger.tex
\begin{figure}[!htbp]
    \centering
    \includegraphics[width=0.45\textwidth]{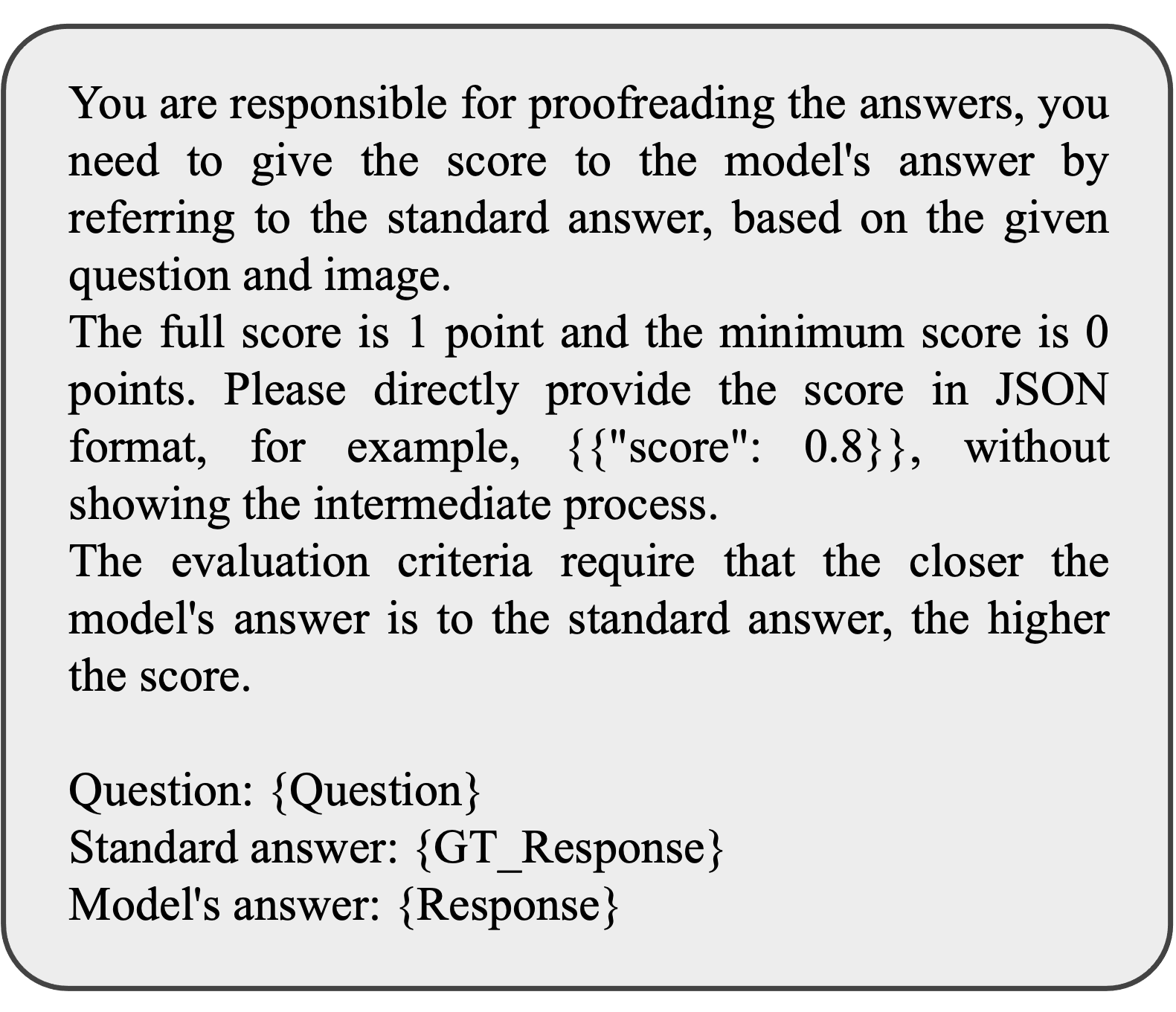}
    \caption{Prompt for the response critics.}\label{fig:res_judger}
\end{figure}

%% file: Figs/demo1.tex
\begin{figure*}[!htbp]
    \centering
    \includegraphics[width=1\textwidth]{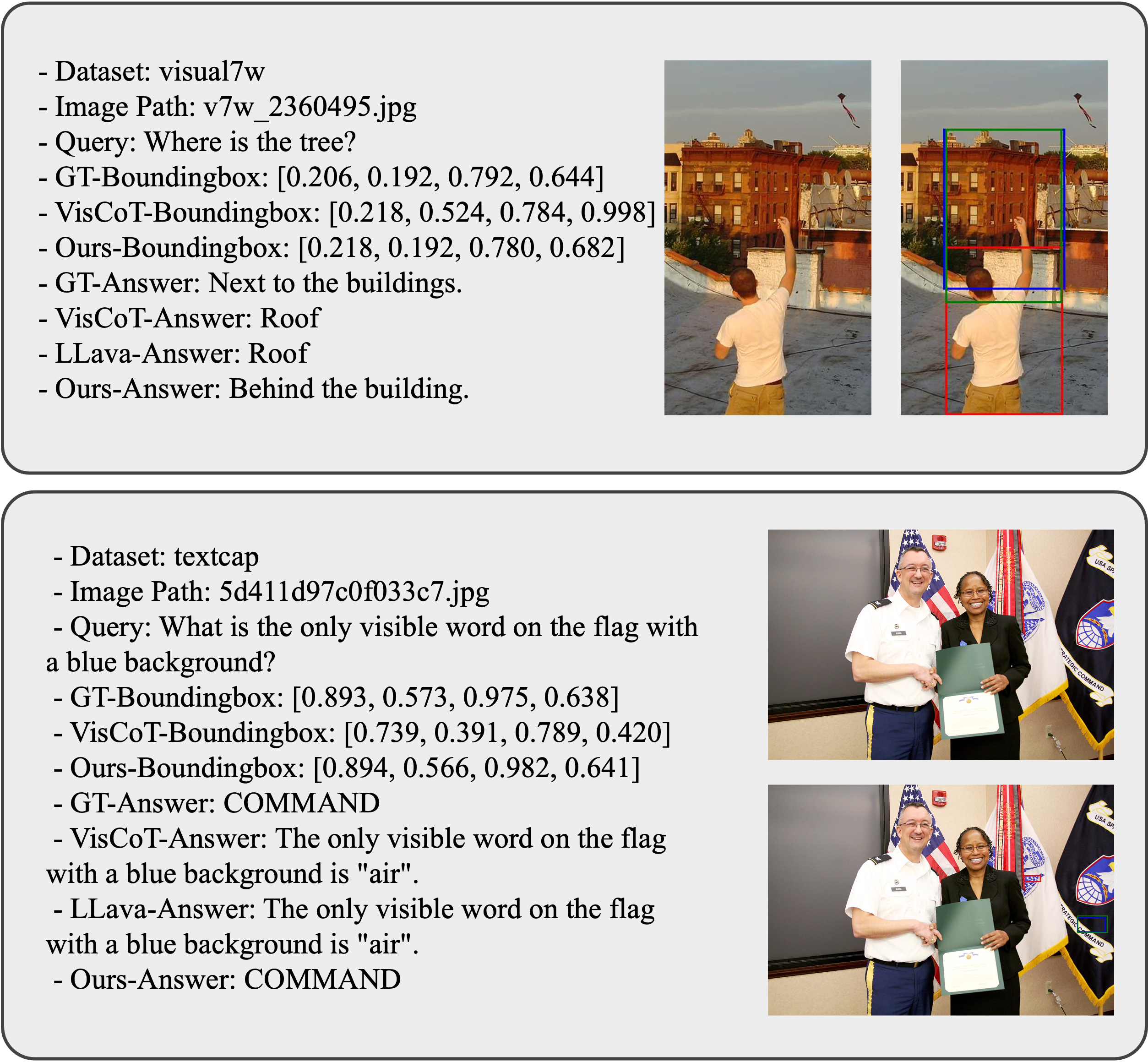}
    \caption{More visualization results of LLaVa-1.5 vs. VisCoT vs. VisRL (based on LLaVa-1.5). Ground truth (GT) bounding boxes are shown in \textcolor{blue}{blue}, VisCoT-generated bounding boxes are shown in \textcolor{red}{red}, while Ours-generated bounding boxes are in \textcolor{green}{green}.}\label{fig:demo1}
\end{figure*}

%% file: Figs/demo2.tex
\begin{figure*}[!htbp]
    \centering
    \includegraphics[width=1\textwidth]{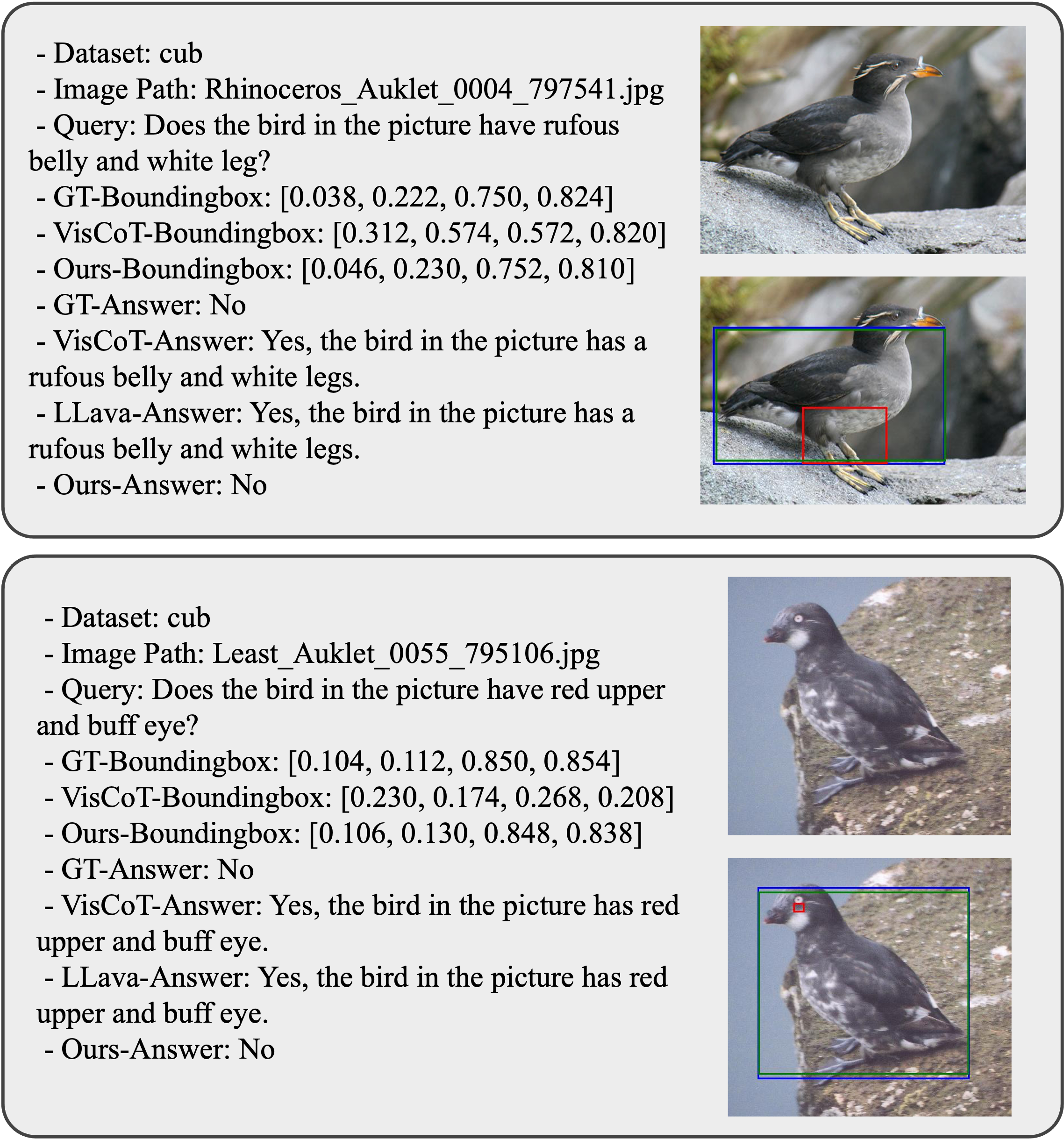}
    \caption{More visualization results of LLaVa-1.5 vs. VisCoT vs. VisRL (based on LLaVa-1.5). Ground truth (GT) bounding boxes are shown in \textcolor{blue}{blue}, VisCoT-generated bounding boxes are shown in \textcolor{red}{red}, while Ours-generated bounding boxes are in \textcolor{green}{green}.}\label{fig:demo2}
\end{figure*}

%% file: Figs/demo3.tex
\begin{figure*}[!htbp]
    \centering
    \includegraphics[width=1\textwidth]{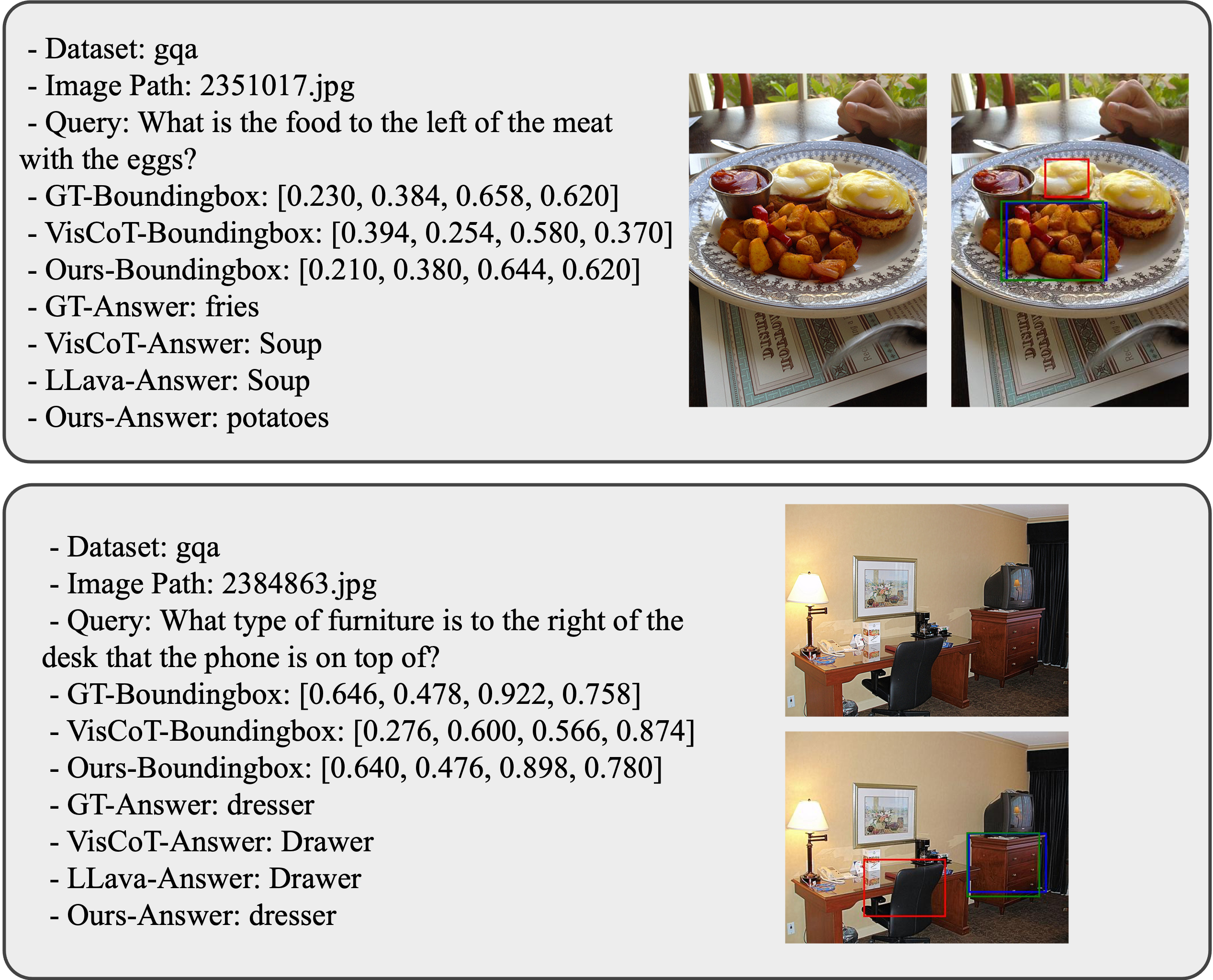}
    \caption{More visualization results of LLaVa-1.5 vs. VisCoT vs. VisRL (based on LLaVa-1.5). Ground truth (GT) bounding boxes are shown in \textcolor{blue}{blue}, VisCoT-generated bounding boxes are shown in \textcolor{red}{red}, while Ours-generated bounding boxes are in \textcolor{green}{green}.}\label{fig:demo3}
\end{figure*}

%% file: Figs/demo4.tex
\begin{figure*}[!htbp]
    \centering
    \includegraphics[width=1\textwidth]{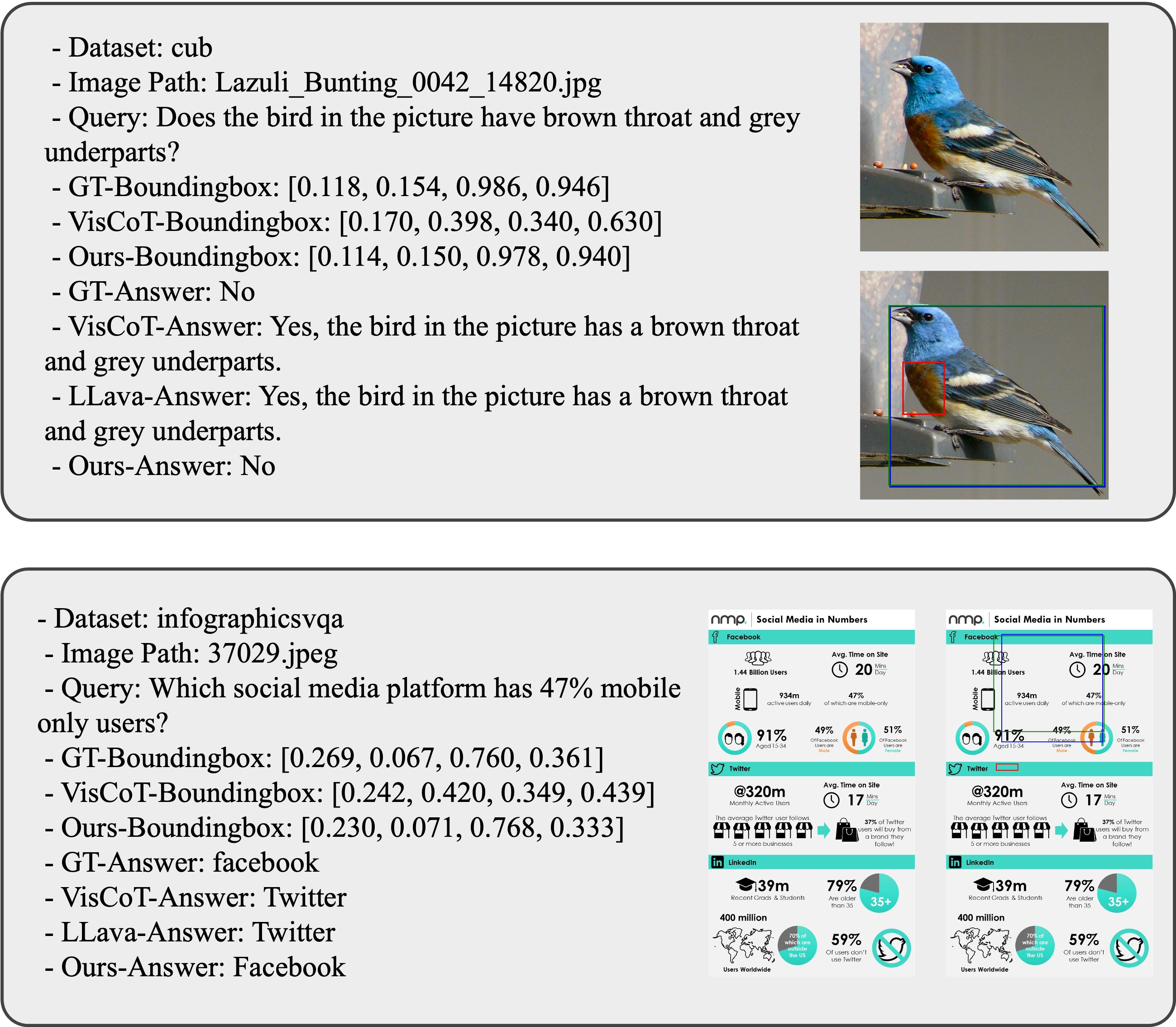}
    \caption{More visualization results of LLaVa-1.5 vs. VisCoT vs. VisRL (based on LLaVa-1.5). Ground truth (GT) bounding boxes are shown in \textcolor{blue}{blue}, VisCoT-generated bounding boxes are shown in \textcolor{red}{red}, while Ours-generated bounding boxes are in \textcolor{green}{green}.}\label{fig:demo4}
\end{figure*}